\documentclass[preprint,12pt]{elsarticle}

\usepackage[english]{babel}
\usepackage[utf8]{inputenc}
\usepackage{amssymb,amsmath}
\usepackage[T1]{fontenc}
\usepackage[title,titletoc]{appendix}
\usepackage{extramarks}
\usepackage{titlesec}
\usepackage{changepage}
\usepackage{mathtools}
\usepackage{float}
\usepackage{color}
\usepackage{makecell}
\usepackage{multicol}
\usepackage[noend, ruled, vlined, linesnumbered]{algorithm2e} % ruled, vlined
\usepackage{listings}
\usepackage{tikz}
\usepackage{environ}
\usepackage{multirow}
\usepackage{caption}
\usepackage{subcaption}
\usepackage{booktabs}
\usepackage{tabularx}
\usepackage{graphicx}
\usepackage{url}
\usepackage[pscoord]{eso-pic}%
\newcommand{\placetextbox}[3]{% \placetextbox{<horizontal pos>}{<vertical pos>}{<stuff>}
  \setbox0=\hbox{#3}% Put <stuff> in a box
  \AddToShipoutPictureFG*{% Add <stuff> to current page foreground
    \put(\LenToUnit{#1\paperwidth},\LenToUnit{#2\paperheight}){\vtop{{\null}\makebox[0pt][c]{#3}}}%
  }%
}%

\definecolor{brick_red_color}{rgb}{0.973, 0.808, 0.80}
\definecolor{brick_blue_color}{rgb}{0.855, 0.91, 0.988}
\definecolor{brick_green_color}{rgb}{0.835, 0.91, 0.831}
\definecolor{brick_orange_color}{rgb}{1.0,0.902,0.8}

\DeclareMathAlphabet{\pazocal}{OMS}{zplm}{m}{n}
\newcommand{\unif}{\pazocal{U}}

\newsavebox\IBox
\NewEnviron{scaletikzpicturetowidth}[1]{%
	\def\tikz@width{#1}%
	\begin{lrbox}{\measure@tikzpicture}%
		\BODY
	\end{lrbox}%
	\pgfmathparse{#1/\wd\measure@tikzpicture}%
	\BODY
}

\usetikzlibrary{shapes,arrows,fit,calc,positioning}

\usepackage{nameref}

\renewcommand{\eqref}[1]{(\ref{#1})}

\newcommand{\rnum}{\mathbb{R}}

\usepackage{url}
\AtBeginDocument{%
  \def\doi#1{\url{https://doi.org/#1}}}
\makeatother

\journal{Automation in Construction}

\begin{document}

\begin{frontmatter}


\title{Metaheuristic planner for cooperative multi-agent wall construction with UAVs}

\cortext[cor1]{Corresponding author}
%%\author[label1]{Basel Elkhapery}
\author[label1]{Basel Elkhapery\corref{cor1}}
\ead{bkhapery@udel.edu}
\author[label2]{Robert Pěnička}
\ead{penicrob@fel.cvut.cz}
\author[label2]{Michal Němec}
\ead{michalnemecdev@gmail.com}
\author[label1]{Mohsin Siddiqui}
\ead{mohsin@udel.edu}

% \affiliation[label1]{organization={Department of Civil \& Environmental Engineering},
%             addressline={University of Delaware}, 
%             city={Newark},
%             postcode={19716}, 
%             state={Delaware},
%             country={USA}}

% \affiliation[label2]{organization={Faculty of Electrical Engineering},
%             addressline={Czech Technical University in Prague, Technicka 2}, 
%             city={Prague},
%             postcode={166 27},
%             country={Czech Republic}}

\address[label1]{{Department of Civil \& Environmental Engineering},
            {University of Delaware}, 
            {Newark},
            {19716}, 
            {Delaware},
            {USA}}

\address[label2]{{Faculty of Electrical Engineering},
            {Czech Technical University in Prague, Technicka 2}, 
            {Prague},
            {166 27},
            {Czech Republic}}

\begin{abstract}
This paper introduces a wall construction planner for Unmanned Aerial Vehicles (UAVs), which uses a Greedy Randomized Adaptive Search Procedure (GRASP) metaheuristic to generate near-time-optimal building plans for even large walls within seconds. This approach addresses one of the most time-consuming and labor-intensive tasks, while also minimizing workers’ safety risks. To achieve this, the wall-building problem is modeled as a variant of the Team Orienteering Problem and is formulated as Mixed-Integer Linear Programming (MILP), with added precedence and concurrence constraints that ensure bricks are built in the correct order and without collision between cooperating agents. The GRASP planner is validated in a realistic simulation and demonstrated to find solutions with similar quality as the optimal MILP, but much faster. Moreover, it outperforms all other state-of-the-art planning approaches in the majority of test cases. This paper presents a significant advancement in the field of automated wall construction, demonstrating the potential of UAVs and optimization algorithms in improving the efficiency and safety of construction projects.

\end{abstract}

\begin{graphicalabstract}

\placetextbox{0.5}{0.93}{
\fbox{
\begin{minipage}{\dimexpr\textwidth-2\fboxsep-2\fboxrule\relax}
Basel Elkhapery, Robert Pěnička, Michal Němec, Mohsin Siddiqui,
\textbf{Metaheuristic planner for cooperative multi-agent wall construction with UAVs}, Automation in Construction, Volume 152, 2023, 104908, ISSN 0926-5805, \url{https://doi.org/10.1016/j.autcon.2023.104908}.
\end{minipage}
}
}%

%% Although a graphical abstract is optional, its use is encouraged as it draws more attention to the online article. The graphical abstract should summarize the contents of the article in a concise, pictorial form designed to capture the attention of a wide readership. Graphical abstracts should be submitted as a separate file in the online submission system. Image size: Please provide an image with a minimum of 531 × 1328 pixels (h × w) or proportionally more. The image should be readable at a size of 5 × 13 cm using a regular screen resolution of 96 dpi. Preferred file types: TIFF, EPS, PDF or MS Office files.
\begin{center}
\includegraphics[width = 0.62\textwidth]{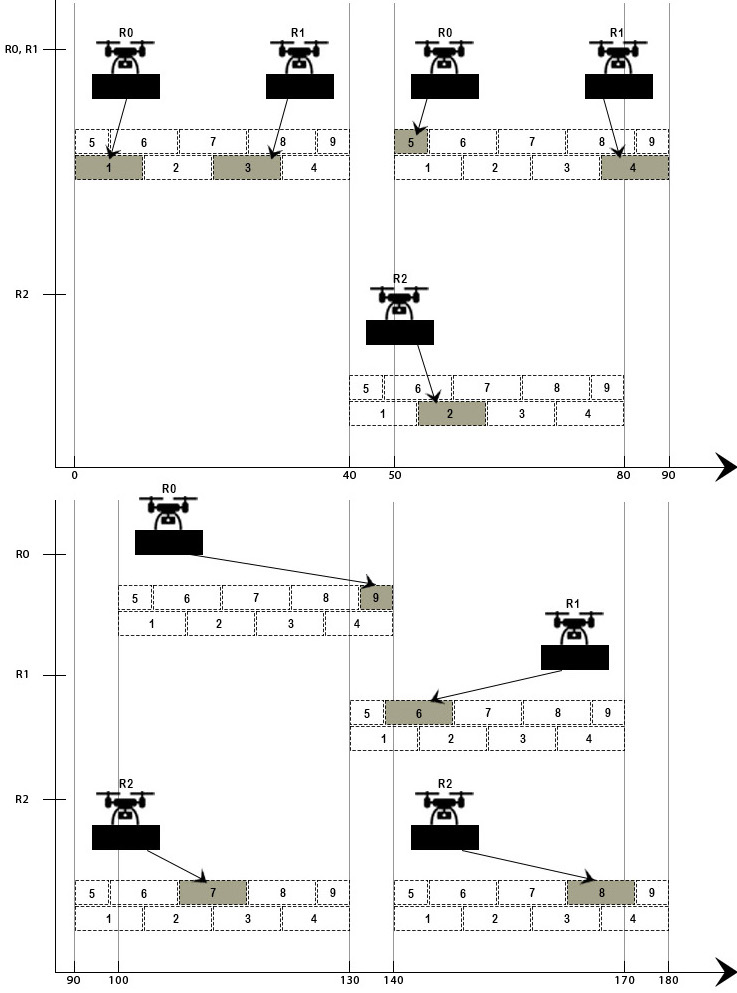}\\
\end{center}
The presented stretcher running wall configuration has multiple feasible assembly solutions, and the designed planner is able to create a near-optimal assembly plan to construct the wall structure while utilizing available unmanned aerial vehicles (UAVs) efficiently. The proposed approach presents an opportunity for a more efficient automated multi-agent construction.

\end{graphicalabstract}


\end{frontmatter}

% \begin{linenumbers}

\section{Introduction}
\label{Introduction}
The global construction industry market is expected to be worth \$15 trillion by 2025, a projected increase of 70\% compared to its worth in 2013 \citep{Vaduva}. Investments in the construction sector are projected to continue to increase steadily shortly beyond 2025 as global economic growth expands. In western nations, this growth is hindered by challenges with labor availability. Accordingly, there is a need for continuous innovation to support and enable the industry to match the rising demand. Furthermore, with the construction industry shift towards high-rise structure development, workers' safety and health risk have undoubtedly escalated. Consequently, construction workers have a 1 in 200 chance of dying or getting severely injured from the job on the field over a 45-year career span period \citep{bigrentz_2022}.

Automation in construction is becoming a prevalent area of research in the fields of computer science and robotics. Construction Automation (CA) and robotic systems development are seen to be able to highly contribute to the development of safer construction, with lower cost, better quality, and reduced duration \citep{PAN2020103174}. Construction robotics can incorporate the latest technologies, such as sensors, cameras, mapping, and planning tools, to allow enhanced control and efficient task execution. Significant research in this area has been done at the Technical University of Munich and throughout Japan and Korea \citep{Vaduva}. Moreover, these techniques tie to the broader field of CA, which is considered an integral technological advancement in developing construction 4.0 practices that promise to revolutionize the industry.

Three families of robotics can be recognized to automate construction activities: 3D printers, mobile robots, and unmanned aerial vehicles (UAVs). Construction 4.0 incorporates all three of these families; however, each has limitations that must still be overcome by research and development. UAVs are currently limited in their accuracy and precision in placing elements. Nevertheless, each iteration of drone advancement incorporates more sensors allowing for safer flights, coupled with software advances that allow for ease of use and autonomy. Furthermore, incorporated technologies have advanced to allow interchangeable and increasing payloads, supporting several construction applications.

Building construction includes many disparate activities, such as site surveying, excavation, bricklaying, concrete casting, tiling, and painting. This paper investigates how the construction of masonry walls can be better automated. Masonry bricklaying is one of the most time-consuming and labor-intensive construction activity. Furthermore, due to the repetitive nature of the processes, bricklaying is seen as an ideal candidate for construction automation. An average mason can place 300-500 bricks a day, whereas a robot can place 800-1200 bricks a day \citep{10.22260/ISARC2017/0140}. CA and robotics can facilitate enhanced construction efficiency with lower costs and allow new work sequences to be realized with the new way of thought resulting from such technological adoption. This research takes the next step forward and builds upon the promising achievements that were carried by the author's \citep{Ba2020AutonomousCW} and other competitors from the Mohamed Bin Zayed International Robotic Challenge (MBZIRC) 2020\footnote{https://www.mbzirc.com/challenge/2020}, challenge number two, where UAVs were required to pick and place bricks on a highlighted wall channel. This research recognizes current UAV limitations; but also recognizes that tools and techniques continue to advance, which will address the precision in future iterations.

Predominantly research on brick construction automation focuses on technological aspects, e.g. attaining required precision and accuracy in placement, and typically sequence bricks deterministically the traditional way ``layer-by-layer''. On the contrary, this work intends to develop a new construction process and enhance construction efficiency by providing a solution in a form of an Orienteering Problem (OP) for sequencing brick construction plan for multiple agents to build a complete wall with a defined bond configuration. The OP is a routing problem, where the objective is to determine the route with limited length through a subset of nodes while collecting the highest score \citep{GUNAWAN2016315}. There are several classes of the OP introduced, including Team OP (TOP), where several agents are utilized to collect the maximum rewards by visiting nodes. Another OP variant is TOP with time windows (TOPTW), where the resources are additionally constrained by a service time window for each node \citep{coptw_paper}.

A brick masonry wall is generally composed of two types of bricks (full-size, and half-size) bricks, but could also include other variants of cut pieces; depending on the size of the wall and the chosen wall bond configuration. The wall-building problem can be classified as a complex task type \citep{taxonomy13}, where there are numerous possible ways to decompose the task of wall-building using multiple agents. The design configuration of a masonry wall allows for a high degree of freedom in a brick placement which allows realizing several viable construction sequencing plans. Reducing the search space with constraints is essential to represent only desired and practical brick placements. Sequencing brick walls is a time-dependent problem where physical constraints can restrict the search space for bricks placement. General constraints for wall building assembly include precedence rules that are directly related to the desired wall bond configuration. Additionally, the proposed algorithm solution must be flexible to be able to incorporate additional placement constraints that arise from utilizing multiple-agents such as UAVs (concurrence rules); to help prevent collision between cooperating agents while constructing the wall.

The wall-building is a combinatorial optimization problem, which can be categorized as a problem with Complex Dependencies~(CD) as defined by~\citep{taxonomy13}. This class of problems relates to task allocation for complex tasks where the objective of an agent depends on the schedules of other agents. The wall-building problem consists of finding an optimal sequence of construction, where we have a certain number of bricks that need to be assigned to a number of available UAV agents to place them on the wall. Each brick assignment requires only one agent and each agent can only place one brick at a time. Therefore, the wall-building problem can be further categorized also by a level 2 designation, defined by \citep{gerkey2004formal} taxonomy as a CD [ST-SR-TA]; where the objective is to compute a time-extended assignment (TA) of single-robot tasks (SR) to single-task robots (ST). Since the robotic agents are constrained with precedence and concurrence placement rules; random placement assignment is not applicable. The choice of subsequent brick placement by other agents depends on the chosen placement of predecessor agents. In other words, an optimal brick placement would be the ones that increase the solution space for subsequent available  agents. A complete algorithm for this problem would need to explore the best decomposition of the problem, and how is the decomposition distributed among agents. The task decomposition problem is connected with the task allocation problem, i.e. optimal task decomposition must be determined concurrently with task allocation.

If a masonry brick wall is composed of low number of bricks, i.e. (5-10 bricks), brute-force algorithms can be implemented and would guarantee finding an optimal sequencing plan by checking all possible solutions. Nevertheless, the number of sequencing solutions scale up permutationally with the increase in number of bricks. Therefore, sequencing wall assembly construction is considered NP-hard \citep{Bulut2015OnTC, PAPADIMITRIOU1984244}. 
Metahuristics methods can be used to find feasible improved solutions for such combinatorial optimization problem in a relatively short time \citep{Gendreau2008}. 
Greedy randomized adaptive search procedure (GRASP) is a metaheursitic method that was first introduced in \citep{FEO198967} as a probabilistic heuristic to solve computationally difficult problems. 
GRASP is an iterative randomized search method in which each iteration provides a feasible solution to combinatorial optimization problems \citep{grasp}. 
The best/final solution after all GRASP iterations is stored as the result. 
Each grasp iteration consists of two phases: 1) construction of a solution via an adaptive randomized greedy function, and 2) local search on the constructed solution to find an improvement.

Other variations for GRASP include Reactive GRASP, where the quality of construction phase iterations are adjusted with improved Restricted Candidate List (RCL) from previous construction phases~\citep{Resende2003}. 
This paper introduces a GRASP metaheuristic planner that can compute a near-optimal plan for multi-agent wall construction collaboration for any desired wall bond configuration. 
The initial feasible solution is formed by greedily selecting elements from the formed RCL. 
RCL elements are generated according to the defined problem space, and entail all the set of nodes (bricks) that could be placed. 
Subsequently, phase two local search is initiated in aims to improve the constructed feasible solution. 
Based on the defined problem formulation, each iteration compares the improved solution with the current best found solution until the defined criterion is met.
The most related versions of the GRASP algorithm are the ones for the variants of TOP in \cite{Expisito2016HeuristicBiasedGRASPTeam,Ruiz-meza2021GRASPSolveMulticonstraints,Sohrabi2020GreedyRandomizedAdaptive,Souffriau2010PathRelinkingApproach,Souffriau2013MulticonstraintTeamOrienteering}.
Yet the proposed variant is mostly based on the vanilla GRASP with RCL introduced in~\cite{Resende2003} and described above.
The major difference to the above versions is given by its application to planning multi-robot wall building.
More specifically, the wall building requires continuous time evaluation, i.e. there is no static graph with task nodes that would be searched for the solution by the GRASP algorithm as in other variants.
Instead, the tasks (individual brick placements) are scheduled on the individual robots' timelines and the GRASP thus has to solve additionally the scheduling problem together with the knapsack and routing problem of the traditional TOP.
Finally, the required precedence and concurrence constraints of brick placement require evaluation in each planning step, which is in contrast to static constraints used for GRASP TOP variants with time windows~\cite{Souffriau2013MulticonstraintTeamOrienteering,Ruiz-meza2021GRASPSolveMulticonstraints}.

\begin{figure}[H]
    \centering
    \includegraphics[width = 1.0\textwidth]{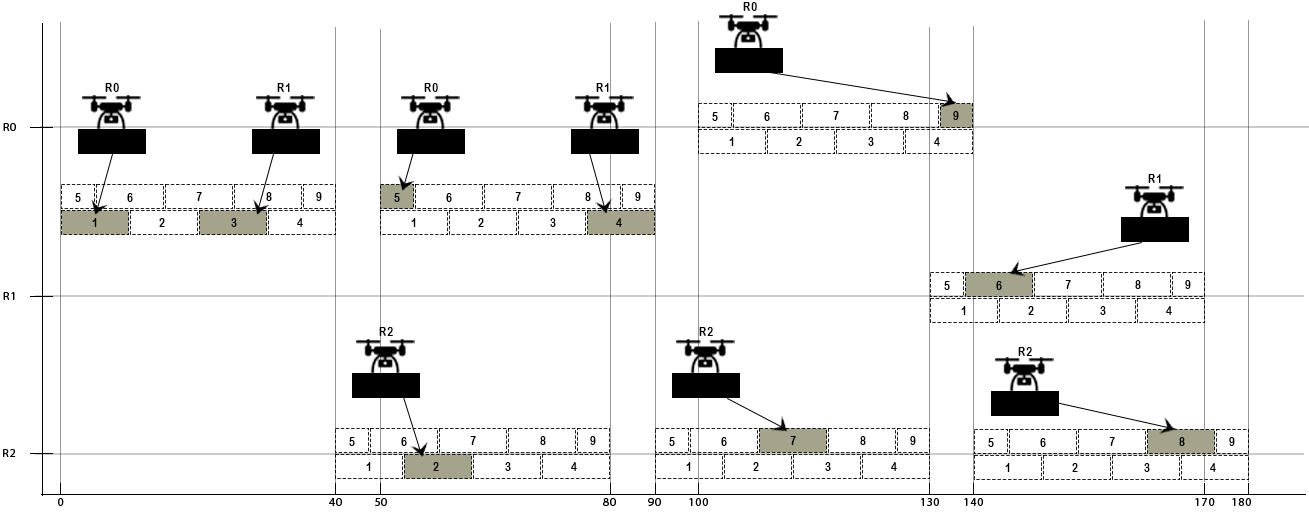}
    \caption{Example two layer wall building sequence plan}
    \label{fig:example_seq}
\end{figure}

In this study we present a complete methodology of automating brick wall construction using multiple agents. Our proposed approach can be applied to plan the sequence of building any wall bond configuration using a heterogeneous team of robots. Lets consider that we are tasked to construct a 2.4-meter wall. The Designer has specified the start and end location of the wall and provided that its of a stretcher running wall bond and 0.4-meter high. Figure \ref{fig:example_seq} portrays the optimal sequence of constructing the wall using 3 agents.

The main problem that this research addresses is how to design a sequence for wall construction using multi-agent robots - given some wall structure (blueprint). However, in attempts to make the process more complete, we also provide the solution of how to design the wall blueprint given its length, height and wall bond configuration. Additionally, we provide an open-source code of our repository\footnote{https://github.com/ctu-mrs/CMCP\_wall\_building\_planner} holding the designed planners and the performed test instances. The research questions are formulated as follows: 1) How can we mathematically formulate the problem of sequencing wall construction? 2) How can we efficiently find a wall building plan for multiple agents such as UAVs? This research addresses multiple gaps in automation of masonry wall construction and in doing so makes important contributions.
First, the study presents an approach that allows multiple agents to simultaneously work on masonry wall construction and even complete the construction of a semi-built wall. 
Second, the proposed planner can develop wall construction plans of even a small wall, that is 16\% percent faster than the classical layer-by-layer approach. 
Third, the proposed planner can compute a near-optimal sequencing plan similar to the exact MILP within a second, while the optimal MILP can not.
Fourth, the proposed planner outperforms all other state-of-the-art planning approaches for wall construction, including GPGP~\citep{Marko2020}, Gurobi optimizer~\citep{gurobi2020gurobi}, and Auction~\citep{nunes2017decentralized}. Lastly, the proposed planner can additionally incorporate the limited flight time of UAVs into the formulation and replenish battery budget with replacement.

\section{Related Work}
\label{Related_Work}
Robots ability to physically interact with the real world; allows them to contribute to construction directly and not just perform inspection applications \citep{Elkhapery_2022, inproceedings}. Previous research recognizes three main directions for applying robotics in automated masonry construction. Some of this research involves using a stationary robotic arm or large-scale robotic structures that can autonomously construct the brick wall or cooperatively work with the mason and assist in lifting operations. The first effort for masonry automation was first attempted in the 1960s – the Motor Mason \citep{pathe}. This machine depended on human masons being around feeding bricks and mortar. Modern technological advancements have allowed other industrial robotics to emerge, including developments by the American Construction Robotics company, like the Material Unit Lift Enhancer (MULE); designed to help masons lift heavy bricks and the Semi-Automated Mason (SAM); utilized for lengthy straight walls \citep{young_2021}. Another significant development that appears to be the most promising of all previous pioneers is the Hadrian X, developed by the Australian Fast brick Robotics; it can construct minor scale building walls autonomously \citep{tokmakova_2020}. Other research utilized UAVs for wall construction, led by Kohler and Raffaello d'Andrea research project - the ``Flight Assembled Architecture'' \citep{Augugliaro2014TheFA}. Other advances succeeding that development include attempts carried out for the Mohamed Bin Zayed International Robotics Challenge (MBZIRC) 2020 \citep{Ba2020AutonomousCW, nvemecplanning}.

As illustrated in Table \ref{tab:related_work}, we further overview the existing task sequencing methods and autonomy level implemented in similar research areas. Most research utilizes blueprints and software, e.g. the architectural planning tool - Grasshopper Rhinoceros\footnote{https://www.grasshopper3d.com}, to sequence brick wall assembly on a layer-by-layer basis. Grasshopper rhinoceros holds all the necessary information about the brick positions and orientation, type of brick – and dependencies, which is directly translated to the robot for execution and construction. Nevertheless, to the best of our knowledge there has been no formulation of an automated nondeterministic construction plan for brick walls.

\begin{table}[H]
    \centering
    \caption{Wall construction task sequencing methods}
    \label{tab:related_work}
    \resizebox{\textwidth}{!}{%
    \begin{tabular}{p{0.15\linewidth}p{0.20\linewidth}p{0.30\linewidth}p{0.30\linewidth}p{0.30\linewidth}}
    \toprule
 Reference & Robot type                 & Structure constructed                                      & Task sequence                                   & Autonomy level                                                                                               \\
    \midrule
\citep{LUPASHIN201441}\\\citep{Augugliaro2014TheFA}& UAVs                       & 6-meter 1500 polystyrene foam brick tower                  & Deterministic plan - from a blueprint           & Human operator to load bricks on the pickup station                                                                                         \\
    \midrule
\citep{FENG2015128}& Mobile robotic manipulator & Prototype brick parametric wall                            & Deterministic layer-by-layer plan - Grasshopper Rhinoceros     & No human interaction                                                                                         \\
    \midrule
\citep{Latteur15}\\\citep{Latteur16}\\\citep{Latteur18}& UAVs                       & Protypes using dricks and droxels                          & Deterministic layer-by-layer plan - from a blueprint               & No human interaction                                                                                         \\
    \midrule
\citep{inbook}& Ground Robot               & Brick parametric wall                                      & Deterministic layer-by-layer plan - Grasshopper Rhinoceros     & Requires human to move the robot every time it finishes a section, in addition to feeding it bricks – 1 type \\
    \midrule
\citep{10.22260/ISARC2017/0140}& Robotic Arm                & Prototype 90 degree wall                                   & Deterministic layer-by-layer plan - BIM(IFC) to KUKA(KRC4)     & No human interaction                                                                                         \\
    \midrule
\citep{Ba2020AutonomousCW}& UAVs and ground robots     & Autonomously placing bricks on a highlighted wall pattern & Deterministic layer-by-layer plan & No human interaction                                                                                         \\
    \midrule
\citep{Marko2020}& UAVs and ground robots     & Autonomously placing bricks on a highlighted wall pattern & GPGP planner algorithm & No human interaction                                                                                         \\
    \midrule
\citep{natureUAV}& UAVs     & 3D printing walls & layer-by-layer deposition & No human interaction                                                                                         \\
    \bottomrule
    \end{tabular}%
    }
\end{table}

Research utilizing UAVs for wall construction tried different ways to overcome the limitation of position imprecisions in elements placement. Studies either modeled parametric walls and used a motion capture system \citep{Augugliaro2014TheFA} or changed the wall construction elements, i.e. dricks (a portmanteau of ``drone'' and ``brick'') and droxels (a portmanteau of ``drone'' and ``voxel'') \citep{Latteur18}. Other research \citep{Ba2020AutonomousCW} utilized a drop channel to place bricks inside, guided by GPS and onboard visual servoing. A later research \citep{natureUAV} presented an augmentation of additive manufacturing and UAVs. Position accuracy limitation was addressed by introducing a dynamic printing head that can improve accuracy to 5mm. In \citep{natureUAV} presented four light cementitious-polymeric composite mixtures that can be carried by the UAVs and harden after material deposition. Nevertheless, such advancements require the development of market forces, and an infrastructure to generate the required material, as opposed to using traditional bricks. In the rest of this section, we briefly discuss these research approaches in automating the wall construction process.

In research led by R. D'Andrea, the Flying Machine Arena (FMA) platform was demonstrated in 2011 at Orleans, France exhibition, where 4 UAVs were utilized to construct a 6-meter-high composed of 1500 polystyrene foam brick tower \citep{Augugliaro2014TheFA}. The space is equipped with 19 Vicon T40 cameras to cover an arena flight space capacity of almost 720 $m^3$ and can record measurements at a rate of at least 300 Hz \citep{LUPASHIN201441}. This experiment is regarded as the first architectural construction assembly with UAVs and is considered the benchmark for future innovations and designs in UAV construction applications. The construction of the structure was manually provided from a blueprint (a plain-text file holding sequential brick placement positions for each UAV).

In \citep{FENG2015128} designed algorithms that automatically determine the sequencing assembly using the designed structure model. Assembly of the wall is constructed the traditional way - layer by layer. The authors used a 7-axis KUKA KR100 robotic manipulator to autonomously assemble several prototype brick parametric wall structures and demonstrate the effectiveness of the designed algorithm. The manipulator was equipped with a camera, and Aruco tag markers were placed on bricks to localize them and to detect current construction progress. Furthermore, a 3D camera was mounted on the manipulator to take 3D point clouds of the structure to help document construction progress.

In aims to present an alternative to additive manufacturing building scale limitation, a seminal research collaboration between MIT and UCL; investigated the feasibility of constructing real building scale structures using UAVs \citep{Latteur16}. Traditional masonry wall construction requires accuracy in the brick assembly of less than 1 cm \citep{Latteur15}. Therefore, the team explored several guidance systems: image recognition and color tags, a laser system mounted on the UAV, GPS coupled with RTK, and GPS coupled with an automatic theodolite instrument. Nevertheless, to overcome UAVs' flight placement imprecision, the study took a different direction by modifying the building's structural components to better fit UAVs' construction limitations. The team used structural elements of different shapes and sizes, including conical-shaped bricks developed to build circular columns \citep{Latteur16}. Additionally, they came up with four families of dricks and droxels. Dricks are geometrically modified bricks with defined grooves that enable overcoming mortar use and accommodate complex and intricate architectural designs, such as curved walls and cantilever \citep{Latteur18}. Both bricks advancements, i.e. designed indentations, guide the bricks to slide in place and allow an inaccurate placement of almost 5cm. Moreover, the team demonstrated CAD benefits and its feasibility in being utilized to formulate automated UAV flight instructions for wall construction on a layer-by-layer basis. For testing purposes, the study implemented manual piloting of a custom-built UAV, with a 40kg payload capacity, to illustrate the pertained tolerance for positioning inaccuracy of the introduced elements in constructing a concrete precast column.

It should be noted that while MIT and UCL team reimagined construction structural elements, introducing such units in the marketplace has significant barriers to implementation. This includes code compliance and building code accommodation, as well as the development of market forces. 

In \citep{inbook} presented an experiment to autonomously construct a dry parametric brick wall  with reduced human intervention. The research used a mobile robot that can detect its location within an indoor lab environment using onboard sensors and developed algorithms. The robot was not designed to navigate autonomously and had to be repositioned 14 times to construct the structure. Accordingly, the designed algorithm computed the sequence from a provided CAD model in a step-wise fashion; to reduce the number of times needed to reposition the robot over the span of the wall. Additionally, to allow for better structure elements alignment, implemented algorithms allowed the robot to compute successor brick positions relative to the placed bricks adaptively. The system was evaluated based on the placed brick position relative to the neighboring bricks on the same layer. It was determined that the placement error was within 3mm of the value expected from the model.

A prototype of a robotic arm was designed by \citep{10.22260/ISARC2017/0140} to construct a prototype 90-degree angled brick wall autonomously. The utilized robotic arm is equipped with a magnetic gripper at its end, and the prototype brick elements are amended with circular magnetic plates on the top. The proposed system uses specially developed software to extract wall elements' positions from a designed BIM model. The designed algorithm forced the robot to abide by principles of a typical mason, e.g. start the construction from the wall corner with a half brick to allow correct linking. Nevertheless, the plan is determined before assembly, and elements are sequenced on a layer-by-layer basis.

In the MBZIRC 2020 competition, challenge 2 consisted of picking and placing bricks by a UAV team on a highlighted wall channel. The best team, ``a collaboration between CTU, UPenn, and NYU'' in the competition, was able to drop ten bricks in the defined wall channel correctly. The presented system \citep{Ba2020AutonomousCW} can scan an area to locate the loading bay and placement location, precisely grasp the bricks, and drop them into the channel. Three UAVs were employed; however, only two were allowed to fly simultaneously to mitigate collisions. Additionally, predicted trajectories were utilized to ensure the UAVs do not collide within their planned routes. The experiment guidance system combined Red-Green-Blue-Depth (RGB-D) camera, GPS, and Light Detection and Ranging (LiDAR) sensors. RGB-D camera was used to precisely locate the brick positions while hovering over the site area. The authors implemented a sequencing planner algorithm using provided blueprints; to place bricks on the wall channel sequentially on a layer-by-layer basis from one side of the wall. Another team \citep{Marko2020} from the competition, proposed a decentralized planning approach to the wall-building mission inspired by Generalized Partial Global Planning (GPGP) \citep{decker1995designing}, with a multi-criteria objective, i.e. quality, cost, and time; in aims to optimize wall construction. The research used 3 different brick sizes and tested their approach on 20 random bond configuration wall datasets with similar dimensionality, made of 7-10 bricks. The authors \citep{Marko2020} employed a heterogeneous team consisting of two UAVs and one UGV on the task of wall-building, and compared their approach with two state-of-the-art literature planning approaches, i.e. Auction \citep{nunes2017decentralized} and Gurobi Optimizer \citep{gurobi2020gurobi}. Results indicate that their approach performs within 12 \% of the optimum.

As highlighted in this section, research on brick construction automation primarily focuses on modifying construction elements and addressing other technological aspects, e.g. attaining required precision and accuracy in placement, and typically sequence bricks deterministically the traditional way ``layer-by-layer''. On the contrary, this work intends to use existing/traditional masonry unit elements and proposes a novel GRASP metaheuristic planner that can take advantage of a multi-agent collaboration for wall construction and compute a near-optimal construction plan in a relatively short time.

\section{Preliminaries - Traditional vs. Robotic wall construction process}
\label{Traditional_vs_Robotic_wall_construction_process}
\begin{figure}[H]
    \centering
    \includegraphics[width = 0.8\textwidth]{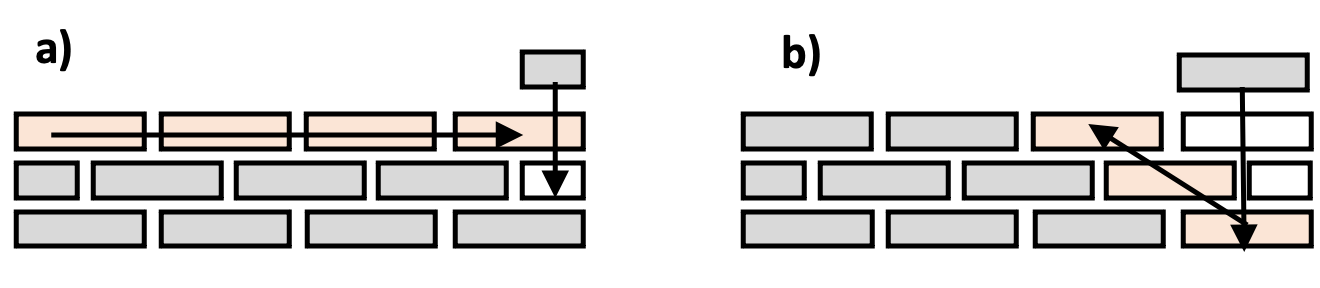}
    \caption{Wall construction methods: a) Traditional b) Stair-wise method}
    \label{fig:method_comp}
\end{figure}

This section describes major differences between the traditional and a general robotic wall construction process. Mason and tools are substituted by a robotic machine (robotic arm with an end-effector or a UAV with a gripper). Secondly, the interface which perceives the design drawing information in the manual process drawings is interpreted by the mason, and the final construction output relies on his experience, understanding of the drawing, and quality of his work. In the robotic process, the final construction output relies on the algorithm that interprets and translates the drawings to information understood by the machine. Thirdly, a robot will not require any additional reference system to aid in the placement of elements apart from the fusion of onboard sensors, e.g. lidar, barometer, camera and RTK GPS. In contrast, a mason typically requires setting up a system of guides to aid in alignment and allow the construction of good-quality walls. Lastly, typical sequencing by a mason is performed the traditional way ``layer-by-layer''; the sequence is either from right to left in every layer or ``vice-versa''. Another unconventional sequencing technique is the stair-wise method ``steps wise fashion'' both methods are presented in Figure \ref{fig:method_comp}.

The traditional way is more predominant as it allows the masons to easily align bricks properly and utilize string lines to help expedite the construction process with lower quality-related issues. Nevertheless, \citep{Vidovszky_2022} compared both methods using a prototype 4-axis robotic arm equipped with a sliding rail, and concluded that the stair-wise method is slightly faster than the traditional method. This is resultant behind the reduced travel time required to reposition the robot over the entire span of the wall in comparison to traditional method construction. This is the represented opportunity that this research tackles pertaining that the quality limitations are overcome with robotic adoption, where the proposed algorithm will determine what order and brick positions to be constructed.

\section{Problem Statement}
\label{Problem_Statement}
A masonry wall is composed of identical elements of a specific size, i.e. bricks. The information needed to construct a wall is typically found in 2D plan design drawings with wall boundaries positions, and dimensions highlighted on a given site layout. Design drawings do not explicitly denote where every brick needs to be placed and do not provide a sequence for placement. However, identifying an optimal plan given the number of resources and current construction progress, is essential for enhanced efficiency and productivity.

\begin{figure}[H]
    \centering
    \includegraphics[width = 0.5\textwidth]{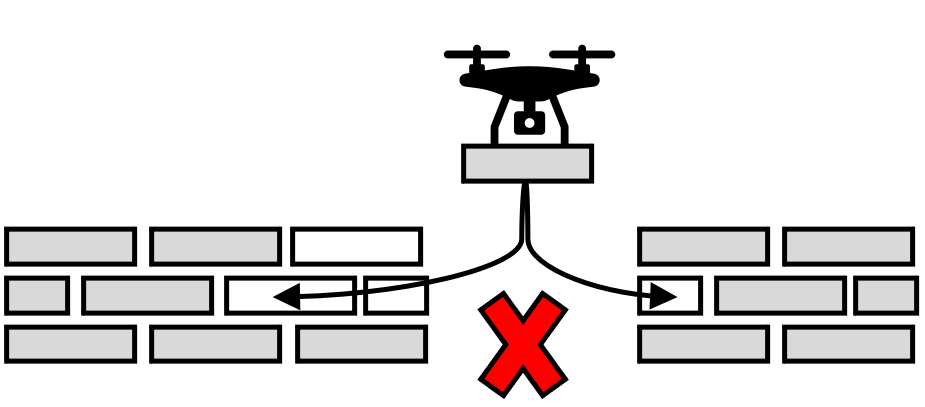}
    \caption{Impractical assembly}
    \label{fig:Impractical}
\end{figure}

As denoted earlier, the brick construction assembly plan can be thought of as a special version of Team Orienteering Problem (TOP), an orienteering problem class, where several resources are utilized to collect the maximum rewards within a specified time budget. In this case, the aim is to maximize rewards by placing bricks within a certain time budget. The time budget essentially ensures that the formulated plan is compact and constrained by the limited battery flight time of a UAV. Additionally, the formulation should incorporate battery replacement of resources to replenish its time budget. Furthermore, the problem should also include building constraints, i.e. precedence and concurrence rules; that prevent any impractical sequencing solutions (Figure \ref{fig:Impractical}). Precedence constraints ensure that bricks are built in the correct order, while concurrence constraints help prevent collision between cooperating agents during construction. A plan is considered feasible as long as it satisfies the defined building constraints.

\subsection{Wall blueprint composition}
\label{Automata_wall_configuration}
In this section we present mathematical rules that describe how bricks come together to form a defined wall bond configuration. These rules are used to generate the blueprint of a straight wall given its length, height and wall bond type. For illustration purposes, we present the rules dictating the blueprint construction of a stretcher running wall bond configuration. The placement of every brick on the wall, from first to last, should not violate any of the below conditions. One can think of rows of bricks as such:

\begin{enumerate}
  \item A brick can come in three forms: complete/full - denoted ``F'', or half-size - denoted ``H'', or empty brick position - denoted ``E''; the ``alphabet'' of bricks is therefore $\sum = \{ F, H, E \}$.
  \item Bricks are concatenated together to form strings.
  \item A wall is a collection of layers ``set of strings''; we say that the wall is partitioned into layers.
  \item Every layer on the wall is indexed $i \in \left<0, \frac{Wall\_height}{Brick\_height}\right>$
  \item n denotes the number of full bricks on a layer i
  \item Each built layer contains a finite string set - denoted ``L'', satisfying the constraint that the concatenation of all the strings in the layer, in any order, and after you erase all occurrences of E, is a sequence $F^k$ with $k < n$.
  \item A string ‘w’ on any built layer can be one of four types:
    \begin{itemize}
        \item $w_i = (F)^k$ : A sequence of $k$ full-size bricks
        \item $w_i = H(F^{n-1})$ : A sequence of a half brick on edge and then full-size bricks
        \item $w_i = F^{n-1}(H)$ : A sequence of full-size bricks and then a half brick on an edge
        \item $w_i = H(F^{n-1})H$ : A sequence of full-size bricks with half-size bricks on each edge
    \end{itemize}
\end{enumerate}

\subsection{TOP problem formulation}
\label{Problem_formulation}
The planning for wall building formulated as TOP is a combinatorial optimization problem, that scales up permutationally with the increase in the number of bricks on the wall. Therefore, it is very challenging to solve it using a generic linear programming solver. Certain rules and assumptions are made to reduce the problem complexity and solution space:

\begin{enumerate}
  \item The defined wall bond configuration is a stretcher-running wall; with bricks on a layer overlap midway with bricks on the layer below and above.  
  \item The wall length is restricted to multiples of a half brick size. This way, it is guaranteed that any brick on the wall can come in only two types/forms: full and half-size bricks.  
  \item All brick’s positions are defined by their center coordinates $\vec{r}_i = \left[\begin{array}{ccc} x_i & y_i & z_i \end{array}\right]$, and orientation around $z$-axis $\varphi_i$.
  \item The $z_i$ coordinate is amended to represent the bottom face of the brick, i.e. a brick on the first layer of the wall would have a $z_i = 0$ and is always an integer multiplication of brick height.
  \item Each brick has an associated reward $\rho_{i}$ based on its attributed surface area, e.g. full-brick = 2, half-brick = 1.
  \item Multiple homogeneous robots are utilized to build the wall cooperatively.
  \item The wall is converted to a special graph representation, with bricks represented as nodes and sequencing relations between bricks represented as edges.
  \item A max time for building the wall is defined to allow modelling of the UAVs limited battery capacity and ensure that the formulated plan is compact.
\end{enumerate}

The proposed Team Orienteering Problem (TOP) model is formulated as a mixed-integer linear program (MILP) with additional placement constraints; to define allowable brick placements and to avoid collision between participating robots during construction, i.e. precedence and concurrence rules. All variables and parameter notations are presented in Table \ref{tab:ild_var}. The decision variables used in the proposed MILP are:
\begin{itemize}
  \itemsep0em 
  \item $x^{r}_{ij}$: binary variable equal to 1 if edge between nodes $v_i$ and $v_j$ for robot $r$ is used, and 0 otherwise;
  \item $z_{ij}$: binary variable equal to 1 if edge between nodes $v_i$ and $v_j$ is used, and 0 otherwise;
  \item $y_i$: binary variable equal to 1 if node $v_i$ is visited, and 0 otherwise;
  \item $s_i$: floating-point variable representing visit time of node $v_{i}$ measured in seconds from start.
\end{itemize}

\begin{table}[H]
    \centering
    \caption{Problem definition variables and parameters}
    \label{tab:ild_var}
	\begin{tabular}{ c | l }
		\hline
		Name & Description \\
		\hline
		$x^{r}_{ij}$ & boolean if edge between nodes $v_{i}$ and $v_{j}$ for robot $r$ is used \\ 
		$z_{ij}$ & boolean if edge between nodes $v_{i}$ and $v_{j}$ is used \\  
		$y_{i}$ & boolean if node $i$ is visited \\    
		$s_{i}$ & visit time of node $v_{i}$ \\
		$S_{i}$ & reward of node $v_{i}$ \\
		$t_{i}$ & duration of node $v_{i}$ \\
		$\beta^{r}$ & start location node of robot r \\
		$V_{f}$ & all non-terminating nodes \\
		$e_{ij}$ & edge between $v_{i}$ and $v_{j}$ \\
		$v_{N}$ & terminating location node\\ 
		$M$ & max time for wall construction \\
		$N$ & number of nodes $N = |V|$\\
        $\mathbf{R} = \{r_i\}$ & set of robots $r_i$ \\
		$\mathbf{V} = \{v_i\}$ &  set of all  nodes \\
		$\mathbf{E} = \{e_{ij}\}$ &  set of all edges between nodes $v_i$ and $v_j$ \\
		$\mathbf{V_f}$ &  set of non-terminating nodes \\
		$\mathbf{\Pi} = \{\pi_i\}$ & set of precedence rules \\
		$\mathbf{\Gamma} = \{\gamma_i\}$ & set of concurrence rules \\
		\hline
	\end{tabular}
\end{table}

Based on the notation introduced in Table \ref{tab:ild_var}, the TOP for the wall-building problem may be formulated as a mixed integer linear program:

\begin{align}
\mbox{Maximize} &\sum_{v_{i} \in V} S_{i} y_{i} - W s_{N}\,, \label{objective_ilp}
\end{align}

Subject to:
\begin{align}
& \sum_{ v_{i} \in V } x^{r}_{{\beta^{r}}\,i} = 1\,,\,\, \forall r \in  \mathbf{R}\,, \label{constraint_start} \\
& \sum_{ r \in  \mathbf{R}}  \sum_{ v_{i} \in \mathbf{V}}  x^{r}_{iN} = |\mathbf{R}| \label{constraint_end}\,, \\
& s_{\beta^{r}} = 0\,,\,\,\forall r \in \mathbf{R} \label{constraint_start_nodes_times}\,, \\
& y_{\beta^{r}} = 1\,,\,\,\forall r \in \mathbf{R} \label{constraint_start_nodes_visited}\,, \\
& y_{N} = 1\,, \label{constraint_end_node_visited} \\
& \sum_{v_{i} \in V \setminus \{v_{c}\}} x^{r}_{ic} = \sum_{v_{i} \in V \setminus \{v_{c}\}} x^{r}_{ci}\,,\,\, \forall v_{c} \in \mathbf{V_f}\,,\,\,\forall r \in  \mathbf{R} \label{constraint_flow_of_robots}\,, \\
& \sum_{ r \in \mathbf{R} } \sum_{ v_{i} \in \mathbf{V}}  x^{r}_{ci} = y_{c}\,,\,\, \forall v_{c} \in \mathbf{V}\,, \label{node_requirements_for_robots} \\
& \sum_{ r \in \mathbf{R} } x^{r}_{ij} \leq |\mathbf{R}|  z_{ij}\,,\,\,\forall e_{ij} \in \mathbf{E}\,, \label{constraint_z_and_x} \\
&  s_{i} +  t_{i} + |e_{ij}| \leq s_{j} + M (1 - z_{ij})\,,\,\,\forall e_{ij} \in \mathbf{E}\,, \label{constraint_start_times_continuity} \\
& y_{b} \geq y_{a}\,,\,\,\forall \pi_{i} \in \mathbf{\Pi} \label{precedence_rules1}\,, \\
& s_{a} \geq s_{b} + y_{b} t_{b}\,,\,\,\forall \pi_{i} \in \mathbf{\Pi} \label{precedence_rules2}\,, \\
& s_{b} \geq s_{a} + y_{a}t_{a} \lor s_{a} \geq s_{b} + y_{b}t_{b} \,,\,\,\forall \gamma_{i} \in \mathbf{\Gamma} \label{concurrence_rules}\,, \\
& 0 \leq s_{i} \leq T_{max}\,,\,\,\forall v_{i} \in \mathbf{V}\,. \label{time_constraint}
\end{align}

The objective function \eqref{objective_ilp} is to maximise the total collected reward while ensuring that the formulated plan is compact and does not contain gaps with no resources utilized. If $T_{max}$ defined in constraint \eqref{time_constraint} is greater than the actual time needed to construct the whole wall, the formulated plan will not be compact and will contain gaps. If $T_{max}$ (optimal time required to construct the wall) is known we can define the normalization constant $W = T_{max}$, to restrict the time optimization function on an interval between $\left<0, 1\right>$. It should be noted that reward is not a compulsory component in the objective function, since the goal of the algorithm is to build ideally the whole wall. Nevertheless, reward maximization is  added to indicate that placing a full brick is a more contributing assignment, in comparison to placing other variants of brick sizes; since it covers a greater surface area of the wall. Constraint \eqref{constraint_start} ensures that each robot starts at a defined location. 
Constraint \eqref{constraint_end} ensures that defined nodes terminate at a final node. 
The initial time for start nodes are set to zero by constraint \eqref{constraint_start_nodes_times}. 
Constraint \eqref{constraint_start_nodes_visited} and \eqref{constraint_end_node_visited} guarantees that the start nodes and the end node are visited respectively. 
The robots' visits flow between all edges is ensured by constraint \eqref{constraint_flow_of_robots}.
Constraint \eqref{node_requirements_for_robots} forces that each visited node has a robot that enters it.
Variables $x^{r}_{ij}$ and $z_{ij}$ are joined together through constraint \eqref{constraint_z_and_x}.
Constraint \eqref{constraint_start_times_continuity} ensures continuity with sufficient travel time allocated for each robot, with constant $M$ defined as $M = T_{max} + \max{t_i} + \max{|e_{ij}|}$. 
We then introduce brick placement constraints; to define allowable brick placements and to avoid collision between participating robots during construction, i.e. precedence and concurrence rules. 
Precedence constraints prevent impractical brick placements. Where $\mathbf{\Pi} = \{\pi_i\,|\,\pi_i = (v_{b},v_{a})\}$ is the set of precedence rules. 
Constraints \eqref{precedence_rules1} and \eqref{precedence_rules2} ensure that node $v_a$ is placed only after $v_b$ is placed, and that sufficient travel time was allocated. Concurrence constraints prevent collision between robots by not allowing simultaneous placements of neighboring bricks. Where $\mathbf{\Gamma} = \{\gamma_i\,|\,\gamma_i = (v_{a},v_{b})\}$ is the set of concurrence rules. Constraint \eqref{concurrence_rules} ensures that nodes $v_{b}$ and $v_{a}$ are not planned to be built at the same time. Finally, time constraint \eqref{time_constraint} is added to optimize our problem solution on a time budget interval $\left<0, T_{max}\right>$.

\subsection{Constraint representation}
\label{Constraint_representation}
\begin{figure}[H]
    \centering
    \includegraphics[width = 1.0\textwidth]{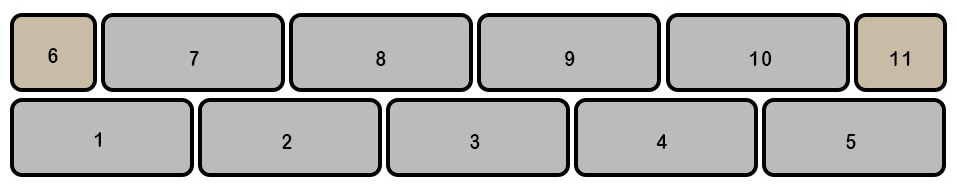}
    \caption{2-layer running brick wall}
    \label{fig:running_wall}
\end{figure}

\begin{figure}[H]
    \centering
    \includegraphics[width = 1.0\textwidth]{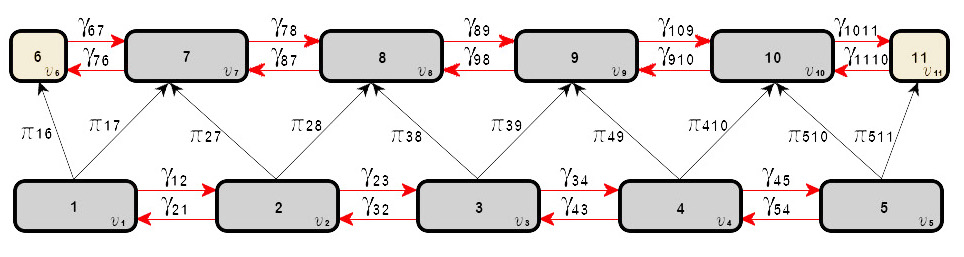}
    \caption{Constraint rules visualization: concurrence (red) and precedence (black)}
    \label{fig:constraint_rep}
\end{figure}

Figure \ref{fig:running_wall} shows an example stretcher running wall bond with two layers of bricks, the first layer is composed of bricks $\{1,2,3,4,5\}$ and the second layer is composed of bricks $\{6,7,8,9,10,11\}$. As defined in MILP formulation, the wall is converted to a special graph representation, illustrated in Figure \ref{fig:constraint_rep}, where bricks are represented by node $v_i \in \mathbf{V}$. Edges $e_{ij}$ are excluded from visualization since they would form a fully connected graph. Precedence rules correspond to $\pi_k$,and concurrence rules correspond to $\gamma_k$. It should be noted that bricks on any layer are only located next to each other at the same elevation and have no physical restriction that prevents them from being placed consecutively in any order. Hence concurrence rules are represented with a double-headed red-colored arrow. However, to help prevent collision between cooperative robots during construction, simultaneous placements of neighboring bricks should be constrained.

\subsubsection{Precedence constraints}
\label{sec:Precedence_rules}
The precedence set $\mathbf{\Pi}$ is generated by inspecting the $z$ component of every brick to establish on which layer each brick resides. Subsequently, a ray casting algorithm for coordinates in 2D $xy$-plane; is utilized to determine which bricks lies on top of another. The precedence set is finally constructed by comparing the top-bottom facing faces of the bricks that were determined to be on top of each other; to check that the bottom-face $xy$-corners are inside the top face.

\subsubsection{Concurrence constraints}
\label{sec:Concurrence_rules}
The concurrence set $\mathbf{\Gamma}$ is generated based on the relative distances between bricks. A distance $d_{min}$ is defined using equation \eqref{d_min}, where $d_{place}$ is a constant that is greater than $\Delta_{ij}$ distance between two adjacent bricks, defined as $d_{ij}$. The concurrence set is finally constructed by selecting bricks that satisfy the performed $d_{min}$ check.

\begin{align}
d_{min} = d_{place}\min_{b_i, b_j \in B}\{|{d}_{ij}|\}\,,\,\, d_{place}\in\rnum\, \label{d_min}
\end{align} 

\subsection{Virtual nodes}
\label{Virtual_nodes}
As presented in our TOP problem formulation (Section \ref{Problem_formulation}), we need start nodes that define start locations for each robot, and an end terminating node as indicated by constraints \eqref{constraint_start} and \eqref{constraint_end}. Accordingly, we introduce virtual nodes which are neither precedence nor concurrence constraints. The start nodes are connected with dotted line edges to all set of bricks that can be built and have no active precedence or concurrence rules $\{1,2,3,4,5\}$. Similarly, the end node is connected with dotted line edges to all set of bricks that the construction plan can end with $\{6,7,8,9,10,11\}$.

\section{Metaheuristic planner}
\label{Metaheuristic_planner}
In this section we present how the proposed Greedy randomized
adaptive search procedure (GRASP) metaheuristic planner can be implemented for multi-agent cooperation to construct any brick wall bond configuration. The plan in essence is constructed in an iterative manner using GRASP to form a feasible solution. Every iteration is composed of multiple sub-routines to determine the set of bricks that are available for placement and have satisfied all the precedence and concurrence rules that were defined earlier (in Sections \ref{sec:Precedence_rules} and \ref{sec:Concurrence_rules}). Subsequently, using local search the feasible solution is improved up on.

The following notations presented in (Table \ref{tab:ild_var1}) are introduced to describe the state of the problem.
\begin{table}[H]
    \centering
    \caption{Additional problem variables}
    \label{tab:ild_var1}
	\begin{tabular}{ c | l }
		\hline
		Name & Description \\
		\hline
		\multicolumn{2}{c}{\textbf{Edges (constraints)}}\\
		\hline
		$s_{ij}$ & visitation time\\ 
		$z_{ij}$ & boolean variable (true if edge was visited, false otherwise)\\
		\hline
		\multicolumn{2}{c}{\textbf{Nodes (bricks)}}\\
		\hline
		$\epsilon_i$ & travel time to reservoir after placing a brick\\ 
		$S_{i}$ & node reward (brick reward)\\
		$t_i$ & time needed to process a node (place a brick)\\
		\hline
		\multicolumn{2}{c}{\textbf{Resources (robots)}}\\
		\hline
		$\eta_i$ & node $\eta_i \in V$ assigned to the resource $\rho_i$ (brick assigned)\\
		$\sigma^s_i$ & start time of node process (brick is picked)\\  
		$\sigma^e_i$ & end time of node process (brick is placed)\\ 
		$\sigma^a_i$ & time needed to process new node (travel to reservoir)\\
		$\lambda_i$ & resources state\\
		$\rho_i$ & number of available IDLE resources\\  
		\hline
		\multicolumn{2}{c}{\textbf{Plan state}}\\
		\hline
		$\mathbf{Z^e_u}$ & set of unvisited edges\\
		$\mathbf{Z^e_v}$ & set of visited edges\\
		$\mathbf{Z^n_u}$ & set of unavailable nodes (set of visited bricks, but constraints not met)\\
		$\mathbf{Z^n_v}$ & set of available nodes (set of bricks that can be placed)\\
		$\mathbf{\Omega}$ & set of already processed nodes (set of placed bricks)\\
		\hline
	\end{tabular}
\end{table}

As illustrated in (Figure \ref{fig:constraint_rep}) in Section \ref{Constraint_representation}, the wall building constraints are illustrated using graph representation. Therefore, we define set of edges $e^{k}_{ij} \in$ $\mathbf{E^{k}}$ for both constraint types. The formed set $\mathbf{E^{k}}$ consists of $\mathbf{E^{p}}$ (denoted $\mathbf{\Pi}$) and $\mathbf{E^{c}}$ (denoted $\mathbf{\Gamma}$), corresponding to precedence and concurrence constraints respectively. The wall is therefore represented on a graph $\mathbf{V, E = \{\bigcup_{k} E^{k} \}}$. Where variable $k$ symbolizes the total number of constraints ``precedence, and concurrence'' identified for a wall. It should be noted that every node on the graph is only considered once all the precedence sets are satisfied. Therefore, we represent each edge $e^k_{ij}$ with two parameters $\{s_{ij}, z_{ij}\}$. Each node $v_i$ has three parameters $\{\epsilon_i, S_{i}, t_i\}$. Each resource (robot) $\rho$ has six parameters $\{\eta_i, \sigma^s_i, \sigma^e_i, \sigma^a_i, \sigma^d_i, \lambda_i\}$. Resources state $\lambda_i$ is described as follows:

\begin{align}
	\lambda_i &= \left\{\begin{array}{ll}
		\mbox{WORKING}, & \text{for } \sigma^s_i\leq t < \sigma^e_i\,,\\
		\mbox{RESTING}, & \text{for } \sigma^e_i\leq t < \sigma^a_i\,, \\
		\mbox{IDLE}, & \text{for } t \geq \sigma^a_i\,.
	\end{array}\right.
\end{align}

Plan states are introduced to keep track of the processed nodes and edges. As presented in Table \ref{tab:ild_var1}, the plan state consists of five sets $\mathbf{\{Z^e_u, Z^e_v, Z^n_u, Z^n_v, \Omega\}}$. Each set is populated with its respective set of edges and nodes during the forward-time evolution process performed by the algorithm to obtain a feasible plan solution. The plan construction starts with all edges stored in set $\mathbf{Z^e_u}$. When an edge is considered, it gets relocated to set $\mathbf{Z^e_v}$. Once a node is determined to have met its constraints and can be placed, it is relocated from set $\mathbf{Z^n_u}$ and is stored in $\mathbf{Z^n_v}$ set. Finally, once a node is processed ``brick is placed'' it gets stored in $\mathbf{\Omega}$ set.

\subsection{Plan construction}
The plan is constructed in an iterative manner using a forward-time evolution process, where the iteration is repeated until one of two conditions is met, either the time constraint is met, or no more nodes are found for assignment. In every iteration step of the plan, all visited edges set $\mathbf{Z^e_v}$ are processed with respect to simulated time $t$. Correspondingly, plan state set $\mathbf{Z^n_v}$ is updated with respective nodes that met the constraints ``bricks that can be placed'' and can be assigned to resources. Lastly, node assignments to resources are performed using a greedy heuristic.

\begin{algorithm}[H]
	\SetAlgoLined
	\KwIn{Real-time plan wall configuration, t, $Z^e_v$, $Z^e_v$, $Z^n_u$, $Z^n_v$}
	process\_edges($Z^e_v$, $Z^e_v$)\; \label{lst:alg:process_edges}
	find\_available\_nodes($Z^n_u$, $Z^n_v$)\; \label{lst:alg:find_available_nodes}
	minimum\_required = assign\_available\_nodes()\;
	\If{no nodes assigned}{
		update\_time(minimum\_required)\;
		\eIf{$t < T_{max}$}{
			place\_assigned\_nodes()\;
		} {
			stop\;
		}
	}
	\caption{Iteration step construction (iterate\_step)}
	\label{alg:iteration_one_step}
\end{algorithm}

The process edges function shown in Algorithm \ref{alg:iteration_one_step} line \ref{lst:alg:process_edges}, finds all visible edges $\mathbf{Z^e_v}$ set at time $t$. Subsequently, for each visited edge $e_{ij}$, we add node $v_j$ to the set of unavailable nodes $\mathbf{Z^n_u}$, unless node $v_j$ is already in the set. Set $\mathbf{Z^n_u}$ will therefore contain the set of bricks that can be assigned to resources once their constraint rules are satisfied. On the other hand, the find available nodes function shown in Algorithm \ref{alg:iteration_one_step} line \ref{lst:alg:find_available_nodes}, uses the set of unavailable nodes $\mathbf{Z^n_u}$ and checks if the edges are among the already found sets for precedence and concurrence rules. Once a node $v_i$ has satisfied all its constraints, it is relocated from unavailable nodes set $\mathbf{Z^n_u}$ to available nodes set $\mathbf{Z^n_v}$.

\subsection{Assign available nodes}
The set of available nodes $\mathbf{Z^n_v}$ is utilized by Algorithm \ref{alg:assing_available_bricks} to determine which nodes are assigned to the resources. Nodes are selected based on greedy assignment with respect to their associated reward (maximum reward preferred). In the case that multiple nodes have similar associated rewards, random assignment is performed. The random assignment is implemented using an integer uniform distribution $\unif(1, |\mbox{Max Reward Nodes}|)$.

\begin{algorithm}[H]
	\SetAlgoLined
	\KwResult{Number of needed resources $\rho_i$}
	max\_reward = 0\;
	max\_reward\_nodes = \{\}\;
	minimum\_resources\_required = $\infty$\;
	\ForEach{$v_i \in Z^n_v$}{
		\If{$\rho_i < \mbox{minimum\_resources\_required}$} {
			minimum\_resources\_required = $\rho_i$\;
		}
		$r_a$ = get\_available\_resources($v_i$)\; \label{lst:alg:get_available_resources}
		\If{$|r_a| \geq \rho_i$} {
			\If{$S_{i}$ $>$ max\_reward} {
				max\_reward = $S_{i}$\;
				reward\_nodes = \{\}\;
			}
			\If{$S_{i}$ = max\_reward} {
				max\_reward\_nodes = max\_reward\_nodes $\cup \left\{v_i\right\}$ \;
			}
		}
	}
		\If{max\_reward\_nodes not empty} {
		$v_p$ = random pick from max\_reward\_nodes\;
		Assign node $v_p$ to available resources\;	
		\textbf{return} 0\;
	}
	\textbf{return} minimum\_resources\_required\;
	\caption{Greedy nodes assignment (assign\_available\_nodes)}
	\label{alg:assing_available_bricks}
\end{algorithm}

The get available resources function shown in Algorithm \ref{alg:assing_available_bricks} line \ref{lst:alg:get_available_resources}, finds the number of available resources based on the actual resources state $\lambda_i$. The algorithm keeps iterating until the desired number of resources $\rho_i$ becomes available ($\lambda_i$ = IDLE).

\subsection{Place assigned nodes}
When an active resource $\rho_i$ is found and assigned a node $\eta_i$, the simulation time $t$ is validated that it is greater than the end time to process the node $\sigma^e_i$. If true, the node is added to the plan, and the respective resource state $\lambda_i$ is changed. The assigned node for placement is designated at time $t$ from $\sigma^s_i$ to $\sigma^e_i$. After the node is placed, precedence edges $e^{p}_{kj}$ are added to the plan state set for unvisited edges $\mathbf{Z^e_u}$ to be processed in the subsequent iteration, whereas concurrence edges $e^{c}_{kj}$ are disabled.

After a resource $\rho_{a}$ is assigned a node $\eta_k = v_i$ to place, the time associated with processing the node (pick, place and travel back to reservoir) is computed. Once all available resources are exhausted (all resources are occupied), that time $t$ is stored to perform a time update procedure, to reflect the time needed for the next resource to be available. The time update procedure is presented in the form of a histogram to identify the minimum time $t$ required for the needed number of resources $\rho$ to become available ($\lambda_i$ = IDLE).

\subsection{GRASP for solution optimization}
As presented in Algorithm \ref{alg:grasp_main} we further improve on solutions found by the greedy search with GRASP optimization procedure. We introduce two variables to define the stopping criterion: the maximum number of iterations $K_{max}$, and the maximum number of iterations where the best solution did not improve $K_{max\,not\,improved}$. While the stopping criterion is not met, phase one of the iteration starts by initializing with a different (Mersenne twister PRNG) seed \citep{10.1145/272991.272995}, to form an initial feasible solution plan; formed by greedily selecting elements from the formed RCL. Subsequently, phase two local search is initiated to improve the constructed feasible solution. Lastly, the actual known best solution is compared with the current found solution. As presented in Algorithm \ref{alg:grasp_main}, reward and completion time $T^{\prime}$ parameters are used to assess the formulated plans. If the plans hold the same rewards, the completion time plan parameter determines which plan is chosen.

\begin{algorithm}[H]
	\SetAlgoLined
	\KwResult{best\_solution}
	\KwIn{found\_solution, best\_solution}
	$k_{iter} = 0$\;
	$k_{not\,improved} = 0$\;
	best\_solution = found\_solution\;
	\While{$k_{iter} < K_{max} \wedge k_{not\,improved} < K_{max\,not\,improved}$}{
		Restart PRNG with different seed\;
		greedy\_solution = greedy\_randomized\_construction(found\_solution)\;
		solution = local\_search(greedy\_solution)\;
		update\_solution(best\_solution, solution)\;
		\If{best\_solution did not improve}{
			$k_{not\,improved} = k_{not\,improved} + 1$\;
		}
		$k_{iter} = k_{iter} + 1$\;
	}
 	\If{found\_solution reward $\geq$ best\_solution reward}{
		\If{found\_solution $T^{\prime}$ $<$ best\_solution $T^{\prime}$}{
			best\_solution = found\_solution\;
		}
	}
	\KwOut{best\_solution}
	\caption{Cooperative Masonry Construction Planner (CMCP) - Greedy Randomized Adaptive Search Procedure}
	\label{alg:grasp_main}
\end{algorithm}

During construction of the plans we save partial plans (defined here as \textbf{snapshots}) to be used later by subsequent construction iterations as an initial solution, as shown in Algorithm \ref{alg:grasp_greedy_randomized_construction}. The defined snapshots encompass all states of the edges, nodes, and resources. The defined initial solution for the algorithm represents both the current state of the wall being constructed and the state of resources operating on the wall. Consequently, greedy randomized construction iterates as defined in Algorithm \ref{alg:iteration_one_step} to return a feasible solution, snapshots, and related states.

\begin{algorithm}[H]
	\SetAlgoLined
	\KwIn{initial\_partial\_solution, global $B^\star_{max}$}
	\KwResult{Feasible solution that satisfies all constraints}
	solution = initial\_partial\_solution\;
	snapshots = \{\}\;
	placed\_nodes = 0\;
	$\Delta_s$ = generate\_snapshot\_positions($B^\star_{max}$)\;
	\While{is not finished} {
		iterate\_step(solution)\;
		\If{solution reward increased} {
			placed\_nodes = placed\_nodes + 1\;
			\If{placed\_nodes $\in \Delta_s$ } {
				snapshots = snapshots $\bigcup$ {solution}\;
			}
		}
	}
	\KwOut{\{solution, snapshots\}}
	\caption{Greedy randomized construction}
	\label{alg:grasp_greedy_randomized_construction}
\end{algorithm}

To reduce the number of snapshots taken and to avoid having memory problems, snapshots are not chosen for every brick placement, instead an adaptive procedure is performed to generate points where a snapshot is to be taken from a previous iteration partial solution. First, we compute $B_{est}$ number of bricks in the found solution, such that $B_{est} \leq B_{max}$, where $B_{max}$ is the total number of bricks on the wall. Algorithm \ref{alg:grasp_greedy_randomized_construction} holds a global state $B^\star_{max}$, the maximum number of bricks placed in the current best solution. $B^\star_{max}$ is updated after the full iteration of the GRASP procedure. Snapshot placement points are generated using set $\Delta_s$ (Equation \ref{eq:Delta_s}) of uniformly chosen non-repeating integer numbers on the interval $\left\{1,B^\star_{max}\right\}$. The size of the set $|\Delta_s|$ is computed using (Equation \ref{eq:Size_delta_s}) with snapshot coefficient $\Upsilon \in \left<0,1\right>$.

\begin{align}
\Delta_s  = \{ \delta_s\,|\,\mbox{non-repeating}\,\delta_s\in \left\{1,B^\star_{max}\right\}\} \label{eq:Delta_s}\,,\\
|\Delta_s|~=~\Upsilon~B^\star_{max} \label{eq:Size_delta_s}\,
\end{align}

\subsubsection{Local search}
As presented in Algorithm \ref{alg:grasp_local_search}, a local search procedure is performed for each snapshot constructed during greedy construction from Algorithm \ref{alg:grasp_greedy_randomized_construction}. This is achieved by using generated snapshots as the starting plan wall configuration used in the forward-time evolution process described earlier. Finally, the local search algorithm explores all stored re-optimized snapshot solutions and returns the best-found solution with the lowest completion time $T^{\prime}$ and highest reward.

\begin{algorithm}[H]
	\SetAlgoLined
	\KwIn{Every snapshot plans}
	final\_solutions = \{\}\;
	\ForEach{snapshot} {
		solution = plan\_construct(snapshot)\;
		final\_solutions =  final\_solutions $\cup$ \{solution\}\;
	}
	best\_found\_solution = solution with lowest completion time $T^{\prime}$ and highest reward from final\_solutions\;
	\KwOut{best\_found\_solution}
	\caption{GRASP - local search (local\_search)}
	\label{alg:grasp_local_search}
\end{algorithm}

\section{Experimental results}
\label{Experimental results}
The proposed planner algorithm is evaluated on an adapted version of the framework designed by the authors \citep{BaaCa2021}, the MRS UAV system, an open-source\footnote{https://github.com/ctu-mrs/mrs\_uav\_system} system. The system utilizes a PixHawk flight controller and encompasses tasks such as state estimation, feedback control, takeoff, and landing. The state machine was integrated and designed using Robotic Operating System\footnote{https://www.ros.org} (ROS) \citep{ros}. The complete design of the state machine is developed on FlexBE Behavior Engine\footnote{http://philserver.bplaced.net/fbe/download.php} software; a high-level state machine that enables users to create complex robot behaviors automatically.

\subsection{System Architecture}
\label{System Architecture}
\begin{figure}[H]
    \centering
	\begin{minipage}[b]{0.49\textwidth}
        \includegraphics[width = 1.0\textwidth]{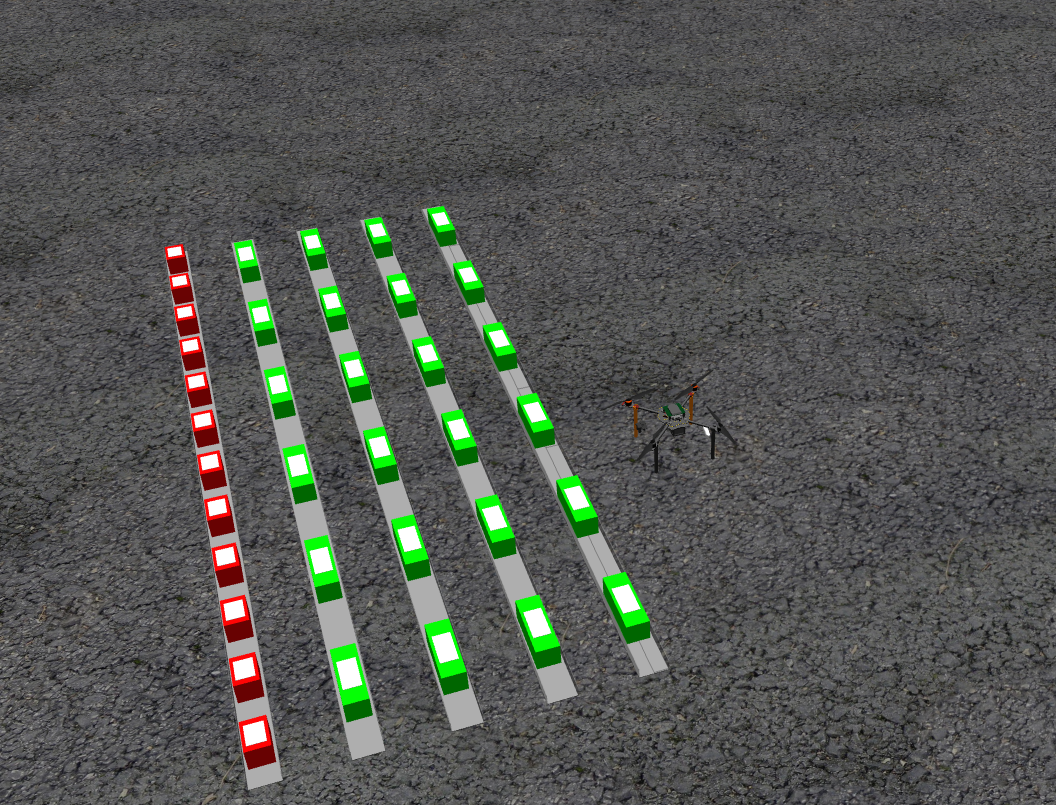}
        \caption{Simulation environment - Brick reservoir}
        \label{fig:simulation}
	\end{minipage}
	\vspace{1cm}
	\hfill
	\begin{minipage}[b]{0.49\textwidth}
        \includegraphics[width = 1.0\textwidth]{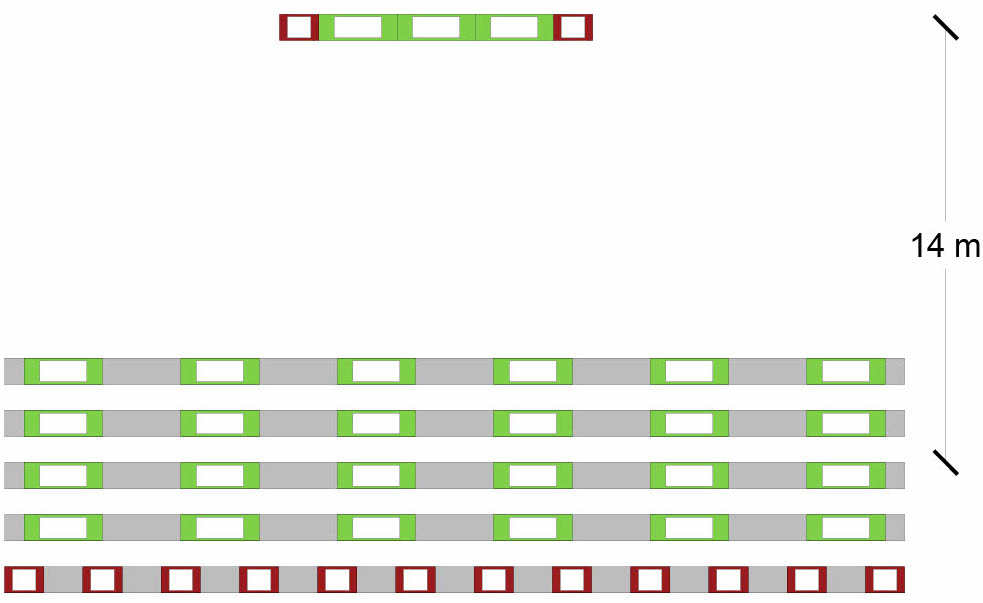}
        \caption{Simulation arena layout}
        \label{fig:Sim_layout}
	\end{minipage}
\end{figure}

The designed planner is tested in gazebo simulator\footnote{https://gazebosim.org/home} environment with ROS. Gazebo 3D simulator allows one to simulate UAVs navigating and interacting with objects (bricks) in a designed environment. The modified system architecture allows UAVs to autonomously construct a brick wall of a specified bond type, position, and length. The generated plan is not deterministic and can change depending on the number of resources assigned to construct the wall. It should be noted that the simulation entails UAVs assembling drywall without using mortar. The guidance system combines RGB-D cameras, GPS, and LiDAR sensors. As presented in Figure \ref{fig:simulation}, the reservoir has four pickup channels with full-size ``green bricks'' and one pickup channel with half-size ``red bricks''. Additionally, UAVs are equipped with a magnetic gripper, and each brick has a ferromagnetic plate attachment on top. Furthermore, as illustrated in Figure \ref{fig:Sim_layout}, the distance between the brick reservoir and the planned wall location is 14 meters.

For the purpose of testing the planner algorithm, we unified the placement duration for all brick types. The performed tests on the gazebo simulation determined that it takes 10 seconds for the UAV to pick a brick from the reservoir $\sigma^s_i$, 20 seconds to place it in the planned wall location $\sigma^e_i$, and $\sigma^a_i =$ 10 seconds travel time back to the reservoir. Accordingly, the total time needed to process a node ``place a brick on the wall location'' is $t_i =$ 40 seconds. An overview of the parameters used by the planner is presented in Table \ref{tab:planner_params}

\begin{table}[H]
    \centering
    \caption{Time needed to process bricks with corresponding rewards}
	\label{tab:planner_params}
	\begin{tabular}{lcc}
		\hline
		& \multicolumn{2}{c}{Bricks} \\
		\cline{2-3}
		parameter   & red & green \\
		\hline
		$S_{i}$   & 1 & 2 \\
		$t_i$   & 40\,s & 40\,s \\
		\hline
	\end{tabular}
\end{table}

\subsection{Wall building State Machine}
\label{Wall building State Machine}
\begin{figure}[H]
    \centering
    \includegraphics[width = 1.0\textwidth]{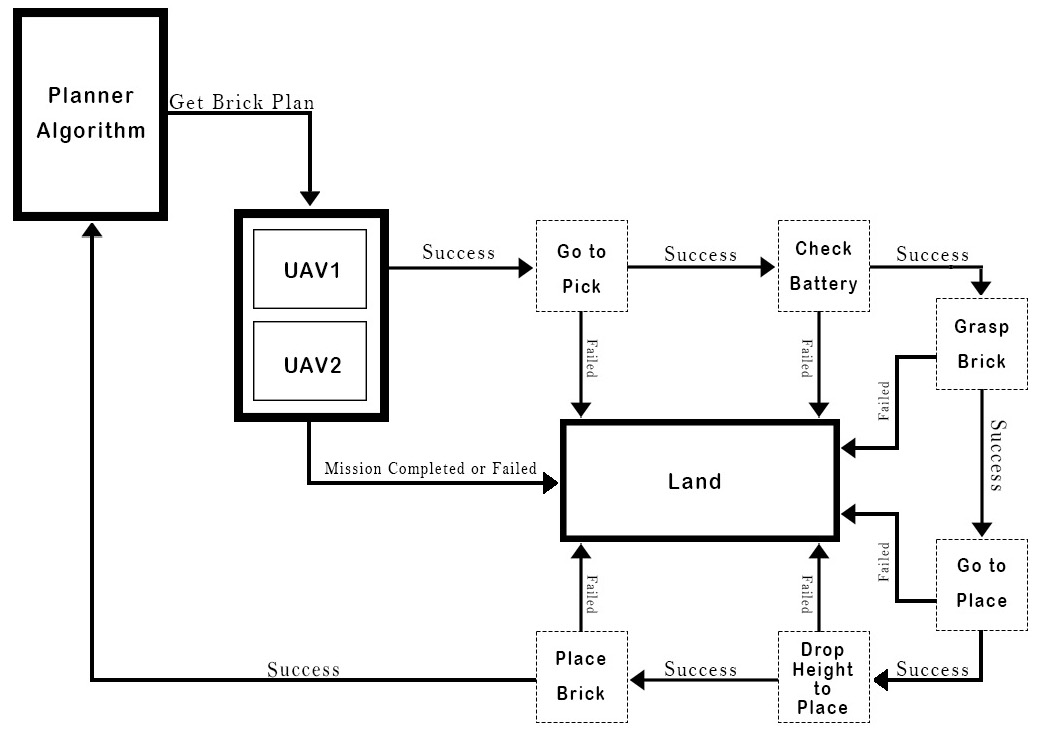}
    \caption{Autonomous Wall Building State Machine}
    \label{fig:state_machine}
\end{figure}

The designed state machine for the UAV autonomous wall-building mission is presented in Figure \ref{fig:state_machine}. The state machine assumes that the UAV is ready for operation and has performed all necessary checks before mission operation, i.e. arming, take off, and fly-in position. Subsequently, our state machine starts by requesting the next brick plan from the designed planner algorithms described in section \ref{Metaheuristic_planner}. The following state, ``Go to Pick'' uses the brick type information and heads to the position of the stack defined in the simulation environment at the height of 3 meters, in which it can see all bricks in the channels and uses the RGB camera to know the brick type and pick the defined type.

\begin{figure}[H]
    \centering
    \includegraphics[width = 0.7\textwidth]{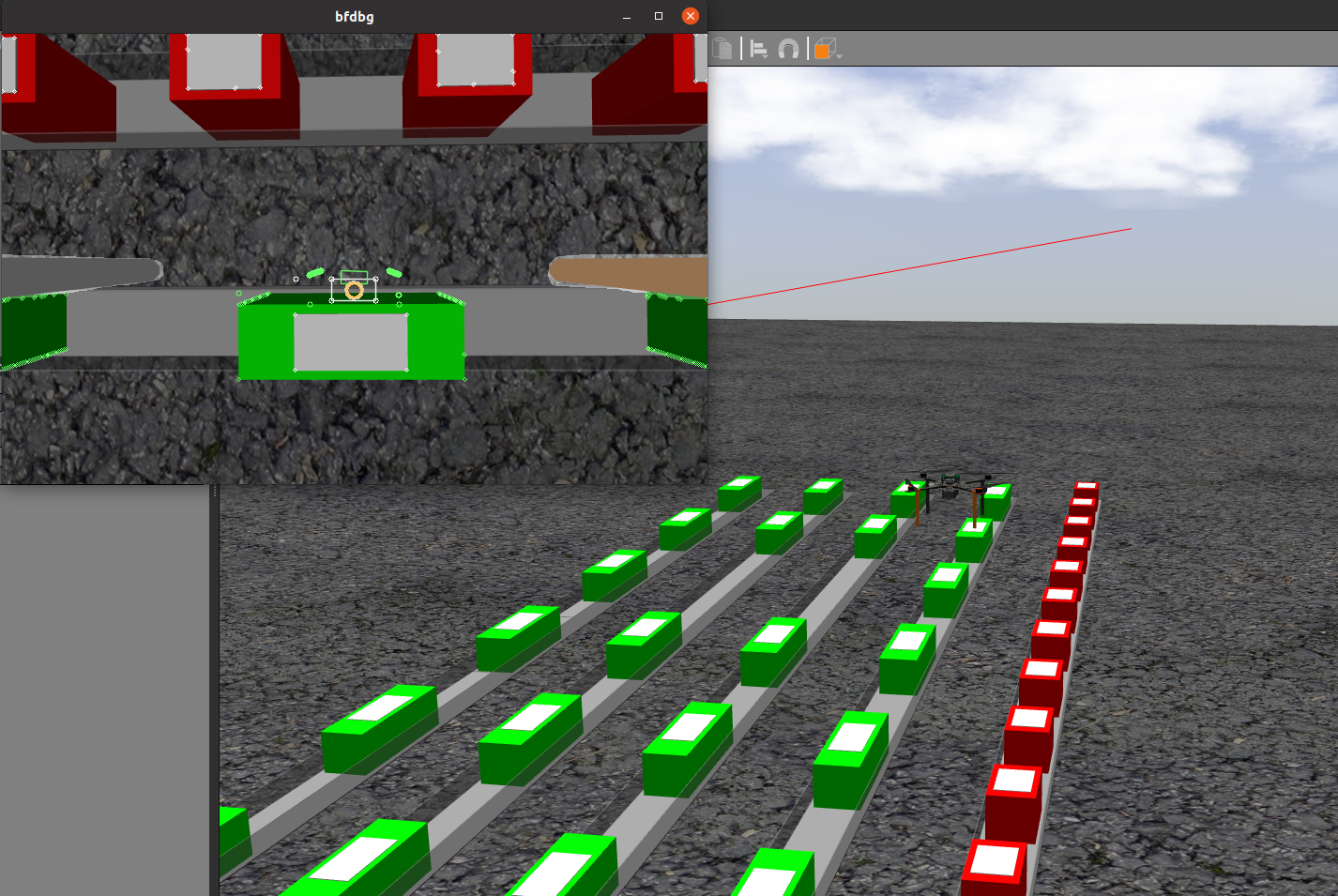}
    \caption{Simulated camera feedback while picking bricks}
    \label{fig:simCameraPick}
\end{figure}

Before grasping operations, the following state machine checks that the UAV battery is sufficient to complete at least one mission cycle and provides a boolean answer to the next state on whether to grasp the brick and continue with the mission or abort and land at home to get the battery replaced. As presented in Figure \ref{fig:simCameraPick}, the UAV uses the camera to align itself over the brick's center location properly and activates the magnetic gripper to attach the brick.

With three failed grasping operations, the state machine of the grasping retries picking a different brick from the pile closer to the one being tried. The ``go to place'' state uses the supplied position coordinates received initially from the get next brick plan and positions itself. The Drop height state requests the drone to descend to the drop height (which is the z coordinate from the position of the brick + 0.2 meters. This drop height was sufficient to place bricks in their location safely. The state machine's last stage is to send a signal to the magnetic gripper to turn off and place the brick before it ascends again. In case there is a placing error, the robot utilizes the same brick assigned earlier and tries to replace it. If the placement succeeds, the next cycle of the state machine starts by requesting the new brick position from the next brick plan service.

\subsection{Performance Evaluation of GRASP, CPLEX, and Naive planners}
\label{PlanEval}
To evaluate the efficiency of the proposed planner; we present two substitutes to the proposed planner method, one plan is generated based on typical traditional masonry construction, and the other plan utilizes IBM ILOG CPLEX\footnote{https://www.ibm.com/analytics/cplex-optimizer} solver. CPLEX is a generic linear programming solver that entails a C++ library that can solve linear-constrained optimization problems using exact algorithms to generate an optimal solution. Nevertheless, as hypothesized earlier, sequencing wall assembly construction is a computationally difficult problem that is NP-hard. Although CPLEX can compute an exact solution for the optimization problem, the drawbacks of such implementation appear as the problem scales up, introducing high computational times and memory usage issues. All the performed tests are run on an Ubuntu Linux system machine with specification and compiler information presented in Table \ref{tab:machine_info}. Note that we used CPLEX version 12.8.0 and we enabled it to run on all 16 threads while the GRASP runs on just a single thread.

\begin{table}[H]
    \centering
    \caption{System information used to run all tests}
	\label{tab:machine_info}
	\begin{tabular}{lr}
		\hline
		System & Ubuntu 20.04.5 LTS \\
		CPU & Intel Core i7-11800H @4.6GHz \\
		RAM & 32GB @2400 MHz \\
		\hline
		Compiler & GCC 9.4.0 \\
		Compilation flags & -std=c++17 -03 \\
		\hline
	\end{tabular}
\end{table}

\begin{figure}[H]
    \centering
    \includegraphics[width = 0.75\textwidth]{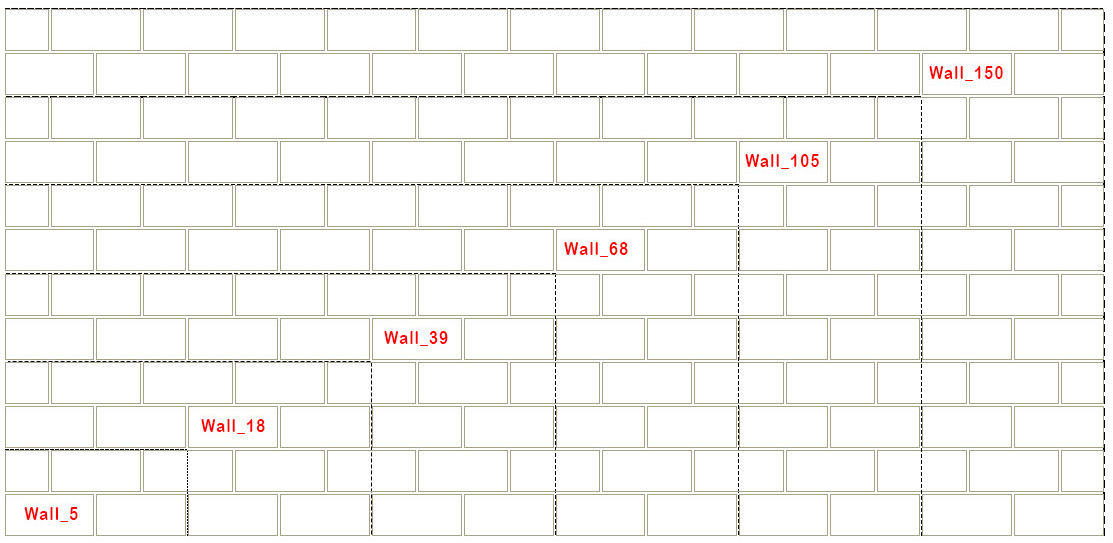}
    \caption{Wall sample dataset}
    \label{fig:wall_sample_dataset}
\end{figure}

To allow comparisons between the proposed GRASP ``Cooperative Masonry Construction Planner'' (CMCP) and the exact solution found by solving functions \eqref{objective_ilp}-\eqref{time_constraint} using CPLEX, we created $6$ different wall datasets with varying sizes (running length and layers), in intervals of plus $2$ as presented in Figure \ref{fig:wall_sample_dataset}.

For all tested scenarios, we restrained the CPLEX solver run time to $2$ hours. Furthermore, we recorded the GAP parameter computed by the CPLEX solver; to highlight the optimality of the found solution. The GAP parameter is defined using equation \eqref{GAP}, where \textit{best bound solution} is an estimate of the best attainable solution, and \textit{best integer solution} is the best solution found by the solver.

\begin{align}
	GAP = 100*\frac{|\mbox{best bound solution} - \mbox{best integer solution}|}{|\mbox{best integer solution}| +1e^{-10}}\,[\%]\, \label{GAP}
\end{align}

On the other hand, for the proposed GRASP planner we defined a snapshot coefficient $\Upsilon = 0.1$ to reduce the number of snapshot creations during random construction and avoid any hardware memory complications. Additionally, since the GRASP planner is stochastic in nature, where results could vary depending on the seed, we have performed 30 runs for each instance and used the estimated average. The results of the tested scenarios for our wall data set samples are presented in Table~\ref{tab:planner_runtime_comparison}. For both planners, we utilized 3 robots/UAV's and had defined $T_{max}=\infty$; to ensure that the complete wall plan is constructed. The numbers defined in the wall dataset names, resemble the number of bricks in the wall. $T^{\prime}$ in Table~\ref{tab:planner_runtime_comparison} denotes the completion time in which the full wall can be built. A smaller $T^{\prime}$ indicates a faster better efficient plan was computed. Progress denotes the percentage completion of the planning progress to build the whole wall, i.e. 100\% progress means that the full construction plan is formed. As expected, results indicate that the CPLEX planner cannot compute an optimal solution; solve the complete wall plan (attain lowest completion time $T^{\prime}$) $GAP = 0.00\%$ within our restrained run time, for walls with more than 18 bricks. On the other hand, our designed CMCP planner was able to efficiently compute a near-optimal solution in less than 1 second runtime for the biggest wall in the dataset wall\_150.

\begin{table}[H]
    \centering
    \caption{Results of CMCP planner vs. CPLEX, 3 robots, $T_{max} = \infty$}
	\label{tab:planner_runtime_comparison}
    \resizebox{\textwidth}{!}{%
	\begin{tabular}{l|rrrr|rrrr}
		\hline
		& \multicolumn{4}{c}{CPLEX} & \multicolumn{4}{c}{CMCP} \\
		\cline{2-9}
		Wall dataset   & progress   & $T^{\prime}$ & runtime & GAP & progress  & $T^{\prime}$ & runtime & $\Upsilon$ \\
		\hline
		wall\_5    & 100\%   & 160\,s   & 44.00\,ms    &  0.00\%     & 100\%   & 160\,s   & 35.69\,ms   & 0.1  \\
		wall\_18   & 100\%   & 360\,s   & 137.09\,ms   & 0.01\%      & 100\%   & 353\,s   & 27.87\,ms   & 0.1  \\
		wall\_39   & 100\%   & 710\,s   & 7200\,s      &  0.39\%     & 100\%   & 680\,s   & 39.49\,ms   & 0.1  \\
		wall\_68   & 100\%   & 1180\,s  & 7200\,s      &  0.37\%     & 100\%   & 1140\,s  & 58.55\,ms   & 0.1  \\
		wall\_105  & 41\%    & 1290\,s  & 7200\,s      &  145.44\%   & 100\%   & 1767\,s  & 377.94\,ms  & 0.1  \\
		wall\_150  & 0\%     & -        & 7200\,s      &  -          & 100\%   & 2490\,s  & 821.45\,ms  & 0.1  \\
		\hline
	\end{tabular}
    }
\end{table}

\begin{figure}[H]
    \centering
    \includegraphics[width = 0.7\textwidth]{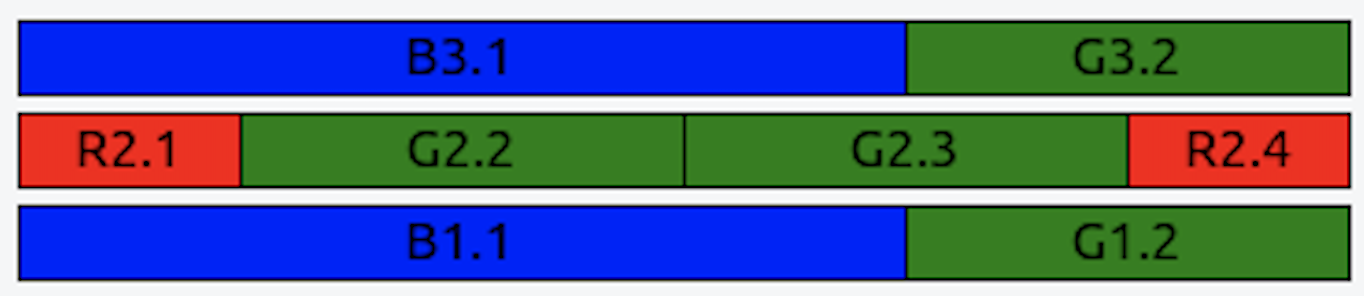}
    \caption{Example Wall from \citep{Marko2020}}
    \label{fig:MarcoWall}
\end{figure}

We further conduct a comparison between our approach and other approaches presented in \citep{Marko2020}, including Generalized Partial Global Planning (GPGP), Gurobi Optimizer \citep{gurobi2020gurobi}, and Auction \citep{nunes2017decentralized}. Similar to CPLEX, Gurobi uses exact methods to solve mixed integer linear programming (MILP) problems and always returns an optimal solution. On the other hand, Auction planning approach uses a single central agent to act as an auctioneer. Table \ref{tab:planner_comparison} presents a comparison of the approaches for 10 wall data sets designed by the authors of~\citep{Marko2020}, with similar sizes but various configurations using three different types of bricks. A sample wall from the dataset is presented in Figure \ref{fig:MarcoWall}.

Similar to \citep{Marko2020}, for this test we implemented a heterogeneous team of robots made of 2 UAVs and 1 UGV to construct each of the walls in the dataset. The UGV is specified under the planner to be slower than the UAV by 10 seconds in placements. Additionally, we matched the placement times for each type of brick. For comparison purposes, we chose to make the comparison with the criteria where only completion time is minimized, as it is the most relevant to our implementation. Table \ref{tab:planner_comparison} shows the comparative analysis between our CMCP planner, CPLEX, and all the approaches presented in \citep{Marko2020}.

\begin{table}[H]
    \centering
    \caption{Completion time $T^{\prime}$ of different planner approaches}
	\label{tab:planner_comparison}
    \resizebox{\textwidth}{!}{%
    \begin{tabular}{l|rr|rr|r|r|r}
		\hline
        \multirow{2}{*}{Planner method} & \multicolumn{2}{c|}{CMCP (ours)} & \multicolumn{2}{c|}{CPLEX (ours)} & GPGP \citep{Marko2020} & Gurobi optimizer \citep{gurobi2020gurobi} & Auction \citep{nunes2017decentralized}\\
        \cline{2-8}
         & $T^{\prime}$ & runtime & $T^{\prime}$ & runtime & $T^{\prime}$ & $T^{\prime}$ & $T^{\prime}$ \\
		\hline
		Set\_1   & 120\,s   & 31.61\,ms    & \textbf{102\,s}    & 106\,ms    & 193\,s    & 111\,s       & 115\,s  \\
		Set\_2   & \textbf{98\,s}    & 35.67\,ms    & 103\,s    & 155\,ms    & 156\,s    & 107\,s       & 103\,s  \\
  		Set\_3   & 109\,s   & 32.56\,ms    & \textbf{107\,s}    & 76\,ms    & 157\,s    & 119\,s       & \textbf{107\,s}  \\  
        Set\_4   & \textbf{94\,s}    & 32.07\,ms    & 102\,s    & 106\,ms    & 158\,s    & 107\,s       & 102\,s  \\  
		Set\_5   & \textbf{80\,s}    & 42.86\,ms    & 119\,s    & 193\,ms    & 158\,s    & 132\,s       & 118\,s  \\
		Set\_6   & 119\,s   & 29.78\,ms    & \textbf{105\,s}    & 192\,ms    & 170\,s    & 125\,s       & 126\,s  \\
		Set\_7   & 103\,s   & 31.54\,ms    & \textbf{96\,s}     & 50\,ms    & 119\,s    & 107\,s       & \textbf{96\,s}  \\
  		Set\_8   & 118\,s   & 35.54\,ms    & \textbf{102\,s}    & 159\,ms    & 178\,s    & 129\,s       & 106\,s  \\  
        Set\_9   & \textbf{87\,s}    & 38.79\,ms    & 96\,s     & 95\,ms    & 144\,s    & 119\,s       & 98\,s  \\  
		Set\_10  & 104\,s   & 29.31\,ms    & \textbf{87\,s}     & 138\,ms    & 133\,s    & 107\,s       & 99\,s  \\   
		\hline 
	\end{tabular}
    }
\end{table}

Results in Table \ref{tab:planner_comparison} highlights the completion time $T^{\prime}$ in which the full wall can be built using available agents in the team. The lowest completion time from each dataset row is highlighed in bold. For our CMCP planner we performed 30 runs for each dataset instance and took the average. As illustrated, our proposed CMCP approach outperforms all other planner approaches in almost half of the test intances. Although, the other half of the instances are surpassed by the CPLEX solution, it does not scale very well with large wall instances as demonstrated earlier. The computation runtime for methods presented in \citep{Marko2020} are not available, however, to further indicate on the significant difference in runtime between CMCP and CPLEX we present the computation runtime for each planner solution. On average, CMCP runtime is almost 4 times faster (385\%) than CPLEX, where its 33\,ms and 127\,ms respectively.

\subsubsection{Evaluation of maximum used resources}
The maximum number of resources that can be assigned to construct a wall depends on several factors, including wall bond configuration, number of layers, and running length of the wall. Similar to traditional masonry construction, given the introduced precedence and concurrence constraint rules that ensure that sequenced bricks are built in the correct order, and prevent collision between cooperating agents during construction; having plenty of resources on a job can be a liability and could even slow down productivity. Note that the proposed model does not prevent agents from going from the start to the end node without laying any bricks, and therefore will have empty plans. Accordingly, we can estimate the maximum number of resources $R_{max}$ that can be utilized by conducting several run tests for the ``number of resources $R$'' using the proposed planner for the same walls dataset developed earlier. As presented in Table \ref{tab:num_resources}, the maximum number of resources that can be used on wall\_18 is 5; any additional resources will remain idle (have an empty plan) since not enough nodes are available due to unmet constraints. Furthermore, as illustrated, adding more than 3 resources to construct the wall does not significantly reduce the completion time $T^{\prime}$. Lastly, results in Table \ref{tab:num_resources} demonstrate how multi-agent planned wall construction contribute to reduced completion time $T^{\prime}$.

\begin{table}[H]
    \centering
    \caption{Assessing maximum number of resources using GRASP}
	\label{tab:num_resources}
    \resizebox{\textwidth}{!}{%
	\begin{tabular}{l|rr|rr|rr|rr|rr}
		\hline
		& \multicolumn{2}{c}{wall\_18} & \multicolumn{2}{c}{wall\_39} & \multicolumn{2}{c}{wall\_68} & \multicolumn{2}{c}{wall\_105} & \multicolumn{2}{c}{wall\_150} \\
		\cline{2-11}
		R   & $R_{used}$ & $T^{\prime}$ & $R_{used}$ & $T^{\prime}$ & $R_{used}$ & $T^{\prime}$ & $R_{used}$ & $T^{\prime}$ & $R_{used}$ & $T^{\prime}$ \\
		\hline
		2    &  2 & 440\,s & 2 & 990\,s & 2 & 1690\,s & 2 & 2640\,s & 2 & 3740\,s\\
		4    &  4 & 340\,s & 4 & 580\,s & 4 & 880\,s & 4 & 1340\,s & 4 & 1890\,s\\
		6    & 5 & 320\,s & 6 & 520\,s & 6 & 700\,s & 6 & 950\,s & 6 & 1300\,s \\
		8    & 5 & 320\,s & 7 & 520\,s & 8 & 710\,s & 8 & 870\,s & 8 & 1080\,s\\
		10   & 5 & 320\,s & 7 & 520\,s & 10 & 680\,s & 10 & 850\,s & 10 & 1050\,s\\
		\hline
	\end{tabular}
    }
\end{table}

Another reason why we allow agents to have empty plans is for re-planning operations, i.e. to allow agents that fails the battery check to return home without having a brick assignment. As illustrated in Table~\ref{tab:planner_runtime_comparison} and Table~\ref{tab:num_resources}, the increase in the scale size of the wall and number of agents; does not significantly impact the computation runtime of our CMCP approach. Moreover, our method scales quite well vertically, i.e. adding more CPU and memory. Accordingly, the proposed method can be used to plan the construction of, e.g., an entire house. However, our current implementation cannot plan for geometric turns, wall corners and other complexities, e.g. window and door openings. Such complexities are planned to be explored in forthcoming research. Additionally, we only looked at dry masonry construction and portrayed the process without the use of mortar. Investigating the coupling and use of traditional cementitious mortars and specialized polymer adhesive is required; to achieve a complete robotic automated construction of wall structures with the required load-carrying capacity and code constraints.

\subsubsection{Plans comparison}
\label{sec:plans_compare}
\begin{figure}[H]
    \centering
    \begin{subfigure}{.5\textwidth}
      \centering
      \includegraphics[width = 1.0\textwidth]{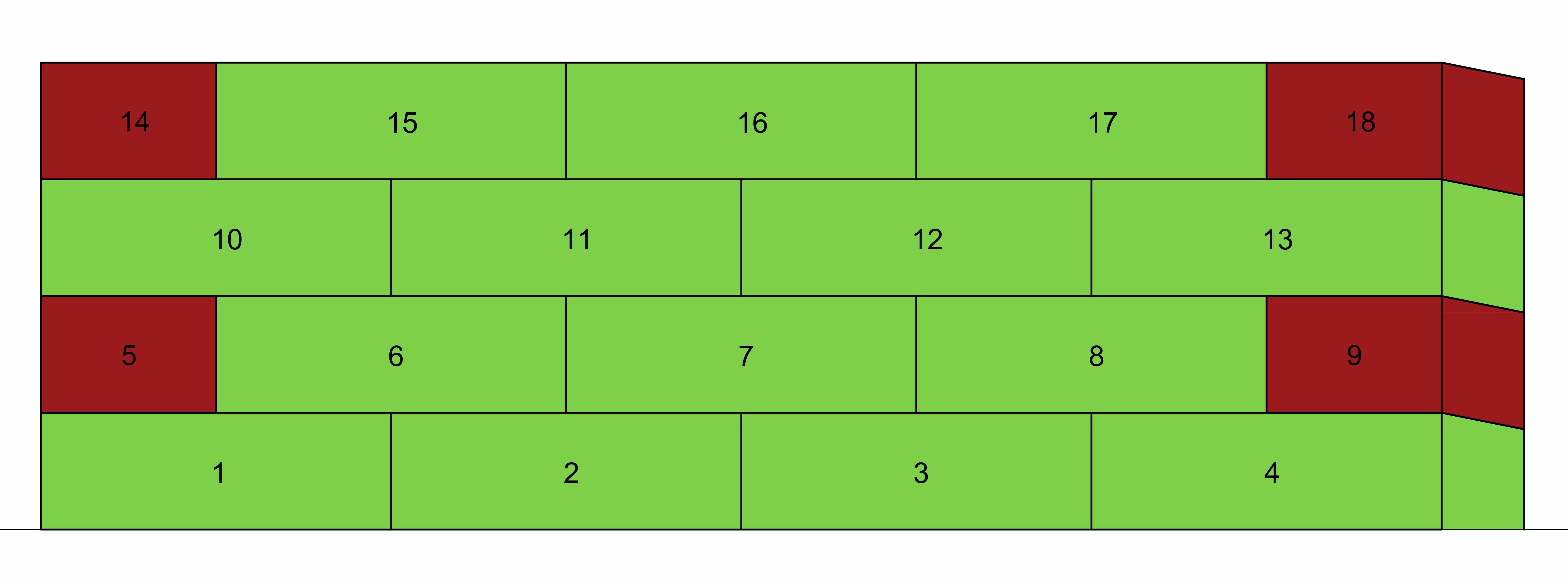}
      \caption{}
      \label{fig:Example_wall}
    \end{subfigure}%
    \begin{subfigure}{.5\textwidth}
      \centering
      \includegraphics[width = 1.0\textwidth]{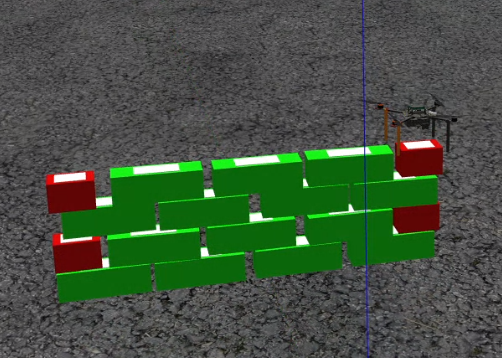}
      \caption{}
      \label{fig:SimFinalWall}
    \end{subfigure}
\caption{(a) Numbered wall\_18 and (b) Simulated built wall\_18}
\label{fig:wall_example}
\end{figure}

Actual solutions obtained from the planner are visually presented using a Gantt chart to highlight the associated actions of each resource at what time period. For representation purposes, the plans comparison is portrayed for wall\_18 dataset presented in Figure \ref{fig:wall_example}, consisting of 18 bricks. The wall bricks are numbered in Figure \ref{fig:Example_wall} from left to right (1 through 18) to enable identification of the selected order of construction with each tested planner. Since the objective of the planner is to build the full wall, the expected maximum collected reward points is $32$, and the optimal plan is selected based on the lowest completion time $T^{\prime}$.

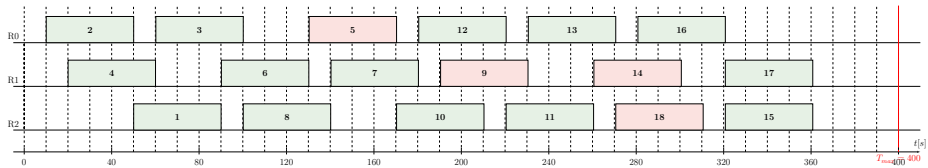
\begin{figure}[H]
	\centering
	\resizebox{0.9\columnwidth}{!}{\begin{tikzpicture}[xscale=1,transform shape]
\draw [-latex](-0.5,0) coordinate(dd)-- (0,0) coordinate (O1) -- (41.00,0)coordinate(ff) node[above]{$t[s]$};
\draw [dashed,thick] (O1) 
-- (0, 5) coordinate(R0) -- (0, 3) coordinate(R1) -- (0, 1) coordinate(R2) -- ++(0,5.7)coordinate(ff2);
\foreach \nn in{R0,R1,R2}{
    \draw [thick] (dd|-\nn) node[above]{\nn}-- (\nn-|ff);
}
\foreach \xx in{1,2,...,40}{
    \draw[dashed] (\xx,0) -- (\xx,0|- ff2);
}
    \draw[thick, color=red] (40,0|- ff2) -- (40,0) node[below]{$T_{max}=400$};
\foreach \xx in{0,4,8,...,40}{
  \ifthenelse{\xx=0}{
      \draw[dashed] (\xx,0.2) -- (\xx,-0.2) node[below]{\xx};
  }{
      \draw[dashed] (\xx,0.2) -- (\xx,-0.2) node[below]{\xx 0};
  }
}
\begin{scope}[shift={(R0)}]
\coordinate(R0) at (1.00,0.6);
\node[ right=0cm and 0cm of R0, right,draw, minimum width=4.00cm,minimum height=1.2cm,fill=brick_green_color!60](b02) {$\mathbf{2}$};
\node[ right=1.00cm of b02, right,draw, minimum width=4.00cm,minimum height=1.2cm,fill=brick_green_color!60](b03) {$\mathbf{3}$};
\node[ right=3.00cm of b03, right,draw, minimum width=4.00cm,minimum height=1.2cm,fill=brick_red_color!60](b05) {$\mathbf{5}$};
\node[ right=1.00cm of b05, right,draw, minimum width=4.00cm,minimum height=1.2cm,fill=brick_green_color!60](b012) {$\mathbf{12}$};
\node[ right=1.00cm of b012, right,draw, minimum width=4.00cm,minimum height=1.2cm,fill=brick_green_color!60](b013) {$\mathbf{13}$};
\node[ right=1.00cm of b013, right,draw, minimum width=4.00cm,minimum height=1.2cm,fill=brick_green_color!60](b016) {$\mathbf{16}$};
\end{scope}
\begin{scope}[shift={(R1)}]
\coordinate(R1) at (2.00,0.6);
\node[ right=0cm and 0cm of R1, right,draw, minimum width=4.00cm,minimum height=1.2cm,fill=brick_green_color!60](b14) {$\mathbf{4}$};
\node[ right=3.00cm of b14, right,draw, minimum width=4.00cm,minimum height=1.2cm,fill=brick_green_color!60](b16) {$\mathbf{6}$};
\node[ right=1.00cm of b16, right,draw, minimum width=4.00cm,minimum height=1.2cm,fill=brick_green_color!60](b17) {$\mathbf{7}$};
\node[ right=1.00cm of b17, right,draw, minimum width=4.00cm,minimum height=1.2cm,fill=brick_red_color!60](b19) {$\mathbf{9}$};
\node[ right=3.00cm of b19, right,draw, minimum width=4.00cm,minimum height=1.2cm,fill=brick_red_color!60](b114) {$\mathbf{14}$};
\node[ right=2.00cm of b114, right,draw, minimum width=4.00cm,minimum height=1.2cm,fill=brick_green_color!60](b117) {$\mathbf{17}$};
\end{scope}
\begin{scope}[shift={(R2)}]
\coordinate(R2) at (5.00,0.6);
\node[ right=0cm and 0cm of R2, right,draw, minimum width=4.00cm,minimum height=1.2cm,fill=brick_green_color!60](b21) {$\mathbf{1}$};
\node[ right=1.00cm of b21, right,draw, minimum width=4.00cm,minimum height=1.2cm,fill=brick_green_color!60](b28) {$\mathbf{8}$};
\node[ right=3.00cm of b28, right,draw, minimum width=4.00cm,minimum height=1.2cm,fill=brick_green_color!60](b210) {$\mathbf{10}$};
\node[ right=1.00cm of b210, right,draw, minimum width=4.00cm,minimum height=1.2cm,fill=brick_green_color!60](b211) {$\mathbf{11}$};
\node[ right=1.00cm of b211, right,draw, minimum width=4.00cm,minimum height=1.2cm,fill=brick_red_color!60](b218) {$\mathbf{18}$};
\node[ right=1.00cm of b218, right,draw, minimum width=4.00cm,minimum height=1.2cm,fill=brick_green_color!60](b215) {$\mathbf{15}$};
\end{scope}
\end{tikzpicture}}
	\caption{CPLEX construction plan}
	\label{fig:plan_cplex_wall_18}
\end{figure}

It should be noted that the defined concurrence rule variable $d_{min}$ for both generated plans was set to $80\,cm$; to avoid collision between cooperating agents during construction. The computed optimal time to construct the example wall as determined by CPLEX is $T^{\prime} = 360\,s$. The corresponding generated sequence plan using 3 robots is presented in Figure \ref{fig:plan_cplex_wall_18}.

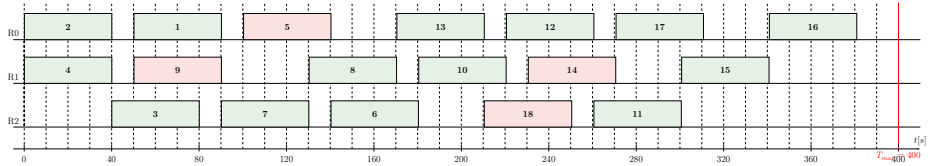
\begin{figure}[H]
	\centering
	\resizebox{0.9\columnwidth}{!}{\begin{tikzpicture}[xscale=1,transform shape]
\draw [-latex](-0.5,0) coordinate(dd)-- (0,0) coordinate (O1) -- (41.00,0)coordinate(ff) node[above]{$t[s]$};
\draw [dashed,thick] (O1) 
-- (0, 5) coordinate(R0) -- (0, 3) coordinate(R1) -- (0, 1) coordinate(R2) -- ++(0,5.7)coordinate(ff2);
\foreach \nn in{R0,R1,R2}{
    \draw [thick] (dd|-\nn) node[above]{\nn}-- (\nn-|ff);
}
\foreach \xx in{1,2,...,40}{
    \draw[dashed] (\xx,0) -- (\xx,0|- ff2);
}
    \draw[thick, color=red] (40,0|- ff2) -- (40,0) node[below]{$T_{max}=400$};
\foreach \xx in{0,4,8,...,40}{
  \ifthenelse{\xx=0}{
      \draw[dashed] (\xx,0.2) -- (\xx,-0.2) node[below]{\xx};
  }{
      \draw[dashed] (\xx,0.2) -- (\xx,-0.2) node[below]{\xx 0};
  }
}
\begin{scope}[shift={(R0)}]
\coordinate(R0) at (0.00,0.6);
\node[ right=0cm and 0cm of R0, right,draw, minimum width=4.00cm,minimum height=1.2cm,fill=brick_green_color!60](b02) {$\mathbf{2}$};
\node[ right=1.00cm of b02, right,draw, minimum width=4.00cm,minimum height=1.2cm,fill=brick_green_color!60](b01) {$\mathbf{1}$};
\node[ right=1.00cm of b01, right,draw, minimum width=4.00cm,minimum height=1.2cm,fill=brick_red_color!60](b05) {$\mathbf{5}$};
\node[ right=3.00cm of b05, right,draw, minimum width=4.00cm,minimum height=1.2cm,fill=brick_green_color!60](b013) {$\mathbf{13}$};
\node[ right=1.00cm of b013, right,draw, minimum width=4.00cm,minimum height=1.2cm,fill=brick_green_color!60](b012) {$\mathbf{12}$};
\node[ right=1.00cm of b012, right,draw, minimum width=4.00cm,minimum height=1.2cm,fill=brick_green_color!60](b017) {$\mathbf{17}$};
\node[ right=3.00cm of b017, right,draw, minimum width=4.00cm,minimum height=1.2cm,fill=brick_green_color!60](b016) {$\mathbf{16}$};
\end{scope}
\begin{scope}[shift={(R1)}]
\coordinate(R1) at (0.00,0.6);
\node[ right=0cm and 0cm of R1, right,draw, minimum width=4.00cm,minimum height=1.2cm,fill=brick_green_color!60](b14) {$\mathbf{4}$};
\node[ right=1.00cm of b14, right,draw, minimum width=4.00cm,minimum height=1.2cm,fill=brick_red_color!60](b19) {$\mathbf{9}$};
\node[ right=4.00cm of b19, right,draw, minimum width=4.00cm,minimum height=1.2cm,fill=brick_green_color!60](b18) {$\mathbf{8}$};
\node[ right=1.00cm of b18, right,draw, minimum width=4.00cm,minimum height=1.2cm,fill=brick_green_color!60](b110) {$\mathbf{10}$};
\node[ right=1.00cm of b110, right,draw, minimum width=4.00cm,minimum height=1.2cm,fill=brick_red_color!60](b114) {$\mathbf{14}$};
\node[ right=3.00cm of b114, right,draw, minimum width=4.00cm,minimum height=1.2cm,fill=brick_green_color!60](b115) {$\mathbf{15}$};
\end{scope}
\begin{scope}[shift={(R2)}]
\coordinate(R2) at (4.00,0.6);
\node[ right=0cm and 0cm of R2, right,draw, minimum width=4.00cm,minimum height=1.2cm,fill=brick_green_color!60](b23) {$\mathbf{3}$};
\node[ right=1.00cm of b23, right,draw, minimum width=4.00cm,minimum height=1.2cm,fill=brick_green_color!60](b27) {$\mathbf{7}$};
\node[ right=1.00cm of b27, right,draw, minimum width=4.00cm,minimum height=1.2cm,fill=brick_green_color!60](b26) {$\mathbf{6}$};
\node[ right=3.00cm of b26, right,draw, minimum width=4.00cm,minimum height=1.2cm,fill=brick_red_color!60](b218) {$\mathbf{18}$};
\node[ right=1.00cm of b218, right,draw, minimum width=4.00cm,minimum height=1.2cm,fill=brick_green_color!60](b211) {$\mathbf{11}$};
\end{scope}
\end{tikzpicture}}
	\caption{GRASP construction plan}
	\label{fig:plan_grasp_wall_18}
\end{figure}

In contrast, for the proposed GRASP method, we performed an extensive search where our $\Upsilon$ parameter for snapshots was set to 1 and maximum defined iterations of $1000$. The performed search took around $45\,\mbox{ms}$ and was able to compute a solution that is close to the optimal plan generated using CPLEX exact method as presented in Figure \ref{fig:plan_grasp_wall_18}. As illustrated, the estimated completion time using our grasp method is $T^{\prime} = 380\,s$.

\begin{figure}[H]
    \centering
    \includegraphics[width = 0.9\textwidth]{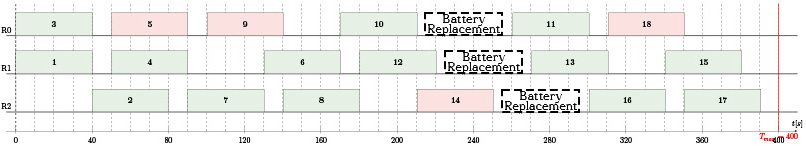}
    \caption{GRASP construction plan with Battery}
    \label{fig:plan_grasp_wall_18_batt}
\end{figure}

For our proposed GRASP method, we additionally included a battery depletion factor into our planner formulation to make the process more complete. When a robot's battery is depleted, the agent is requested to land at home coordinates ``as presented in the state machine Figure \ref{fig:state_machine}''; where the battery of the robot is replaced and made ready again for operation. The battery depletion budget for each robot is defined as 200 seconds, whereas the allocated time for the replacement process was determined to be around 40 seconds as determined from conducted real lab test experiments. The complete construction plan of example wall\_18 with battery replacement process is illustrated in Figure \ref{fig:plan_grasp_wall_18_batt}. 

It should be noted that although the assigned battery depletion variable was set to 200 seconds, the battery depletion time is computed with an offset to the maximum battery life. In other words, if a brick is assigned to a robot before it is depleted, the robot is still able to complete the current cycle before getting the battery replaced. This is illustrated in Figure \ref{fig:plan_grasp_wall_18_batt}, at times 210 and 220 with Robot 0 and Robot 1. The adjusted completion time to construct the example wall as determined by our GRASP method with battery depletion implementation is $T^{\prime} = 390\,s$.

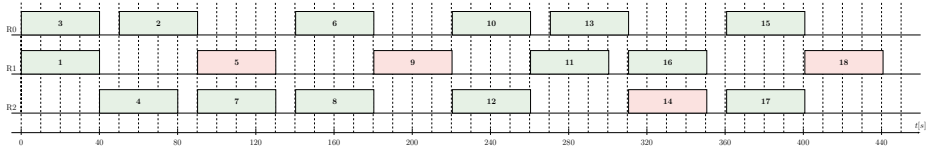
\begin{figure}[H]
	\centering
	\resizebox{0.9\columnwidth}{!}{\begin{tikzpicture}[xscale=1,transform shape]
\draw [-latex](-0.5,0) coordinate(dd)-- (0,0) coordinate (O1) -- (46.00,0)coordinate(ff) node[above]{$t[s]$};
\draw [dashed,thick] (O1) 
-- (0, 5) coordinate(R0) -- (0, 3) coordinate(R1) -- (0, 1) coordinate(R2) -- ++(0,5.7)coordinate(ff2);
\foreach \nn in{R0,R1,R2}{
    \draw [thick] (dd|-\nn) node[above]{\nn}-- (\nn-|ff);
}
\foreach \xx in{1,2,...,45}{
    \draw[dashed] (\xx,0) -- (\xx,0|- ff2);
}
    
\foreach \xx in{0,4,8,...,45}{
  \ifthenelse{\xx=0}{
      \draw[dashed] (\xx,0.2) -- (\xx,-0.2) node[below]{\xx};
  }{
      \draw[dashed] (\xx,0.2) -- (\xx,-0.2) node[below]{\xx 0};
  }
}
\begin{scope}[shift={(R0)}]
\coordinate(R0) at (0.00,0.6);
\node[ right=0cm and 0cm of R0, right,draw, minimum width=4.00cm,minimum height=1.2cm,fill=brick_green_color!60](b03) {$\mathbf{3}$};
\node[ right=1.00cm of b03, right,draw, minimum width=4.00cm,minimum height=1.2cm,fill=brick_green_color!60](b02) {$\mathbf{2}$};
\node[ right=5.00cm of b02, right,draw, minimum width=4.00cm,minimum height=1.2cm,fill=brick_green_color!60](b06) {$\mathbf{6}$};
\node[ right=4.00cm of b06, right,draw, minimum width=4.00cm,minimum height=1.2cm,fill=brick_green_color!60](b010) {$\mathbf{10}$};
\node[ right=1.00cm of b010, right,draw, minimum width=4.00cm,minimum height=1.2cm,fill=brick_green_color!60](b013) {$\mathbf{13}$};
\node[ right=5.00cm of b013, right,draw, minimum width=4.00cm,minimum height=1.2cm,fill=brick_green_color!60](b015) {$\mathbf{15}$};
\end{scope}
\begin{scope}[shift={(R1)}]
\coordinate(R1) at (0.00,0.6);
\node[ right=0cm and 0cm of R1, right,draw, minimum width=4.00cm,minimum height=1.2cm,fill=brick_green_color!60](b11) {$\mathbf{1}$};
\node[ right=5.00cm of b11, right,draw, minimum width=4.00cm,minimum height=1.2cm,fill=brick_red_color!60](b15) {$\mathbf{5}$};
\node[ right=5.00cm of b15, right,draw, minimum width=4.00cm,minimum height=1.2cm,fill=brick_red_color!60](b19) {$\mathbf{9}$};
\node[ right=4.00cm of b19, right,draw, minimum width=4.00cm,minimum height=1.2cm,fill=brick_green_color!60](b111) {$\mathbf{11}$};
\node[ right=1.00cm of b111, right,draw, minimum width=4.00cm,minimum height=1.2cm,fill=brick_green_color!60](b116) {$\mathbf{16}$};
\node[ right=5.00cm of b116, right,draw, minimum width=4.00cm,minimum height=1.2cm,fill=brick_red_color!60](b118) {$\mathbf{18}$};
\end{scope}
\begin{scope}[shift={(R2)}]
\coordinate(R2) at (4.00,0.6);
\node[ right=0cm and 0cm of R2, right,draw, minimum width=4.00cm,minimum height=1.2cm,fill=brick_green_color!60](b24) {$\mathbf{4}$};
\node[ right=1.00cm of b24, right,draw, minimum width=4.00cm,minimum height=1.2cm,fill=brick_green_color!60](b27) {$\mathbf{7}$};
\node[ right=1.00cm of b27, right,draw, minimum width=4.00cm,minimum height=1.2cm,fill=brick_green_color!60](b28) {$\mathbf{8}$};
\node[ right=4.00cm of b28, right,draw, minimum width=4.00cm,minimum height=1.2cm,fill=brick_green_color!60](b212) {$\mathbf{12}$};
\node[ right=5.00cm of b212, right,draw, minimum width=4.00cm,minimum height=1.2cm,fill=brick_red_color!60](b214) {$\mathbf{14}$};
\node[ right=1.00cm of b214, right,draw, minimum width=4.00cm,minimum height=1.2cm,fill=brick_green_color!60](b217) {$\mathbf{17}$};
\end{scope}
\end{tikzpicture}}
	\caption{Traditional construction plan}
	\label{fig:plan_trad_wall_18}
\end{figure}

Finally, we also compare the GRASP planner with traditional naive planner; that builds the wall in a layer-wise fashion as typical traditional masonry would be performed. Additionally, the planner is constrained with concurrence rules to allow for sufficient working space for agents constructing the wall. As presented in Figure \ref{fig:plan_trad_wall_18}, the traditional plan method attained a completion time $T^{\prime} = 440\,s$, which is clearly not efficient as the proposed approach.

\section{Conclusions}
\label{Conclusion}
This paper describes an algorithm for multi-agent automated wall construction called CMCP (Cooperative Masonry Construction Planner) with the aim of enhancing workers' safety, improving quality, and increasing productivity. The problem is formulated as mixed-integer linear programming with added wall-building, precedence and concurrence rule constraints, that ensure bricks are built in the correct order and help prevent collision between cooperating agents during construction. A methodology using drones to construct masonry walls as a case scenario was outlined, with salient elements of the approach discussed. The approach utilized a metaheuristic GRASP planner that computes a near-optimal sequencing arrangement for the structure to be built. The proposed planner computation time was significantly lower than the exact method CPLEX planner, where the biggest wall sample ``wall\_150'' had a computation runtime of 0.82\,s. On the other hand, CPLEX planner was unable to compute a feasible solution plan in less than 2 hours for walls with more than 100 bricks. Moreover, the proposed CMCP planner outperforms in the majority of the test cases all other state-of-the-art planning approaches for wall construction, including GPGP, Gurobi optimizer and Auction. Lastly, the CMCP planner approach presented a better efficient construction plan with battery depletion incorporated as opposed to traditional construction plan, where the completion time $T^{\prime}$ for ``wall\_18'' was $390\,s$ and $440\,s$ respectively.

The current implementation of our method is limited to dry masonry construction where the process is portrayed without the use of mortar. Investigating the coupling and use of traditional cementitious mortars and specialized polymer adhesive is required; to achieve a complete robotic automated construction of wall structures. Additionally, the proposed planner cannot plan for geometric turns, wall corners, and other complexities, e.g. window and door openings. Such complexities are planned to be explored in forthcoming research. The authors envision that with further development of the current framework, we can enhance wall construction productivity and take advantage of mass-produced bricks by utilizing a heterogeneous team of robots to construct an entire house. For example, a team would include brick-laying robots, mortar-placing robots, and 3d-printing robots to print complexities such as corners, doors, and windows.

\section{Acknowledgements}
\label{Acknowledgements}
We would like to thank the MRS\_UAV team for their intricate system design which we adapted to simulate our proposed planner approach. We would also like to thank Dr. Tanner Herbert for his support and valuable insights to making this work successful. The presented work has been supported by the Czech Science Foundation (GAČR) under research project No. 23-06162M.

% \end{linenumbers}

\vspace{1.0em}

\vspace{-1.5em}
\section*{Supplementary Material}
{\small
\vspace{-0.1em}
\noindent \textbf{Code:} \url{https://github.com/ctu-mrs/CMCP_wall_building_planner}
}

%%       \citet{<label>} ==> Jones et al. [21]
%%       \citep{<label>} ==> [21]

% \bibliographystyle{plain} 
\bibliographystyle{elsarticle-harv} 
\bibliography{BibJournal.bib}
% \printbibliography

\end{document}